\documentclass[letterpaper, 10pt, conference]{ieeeconf}

\IEEEoverridecommandlockouts
\let\labelindent\relax

\usepackage{graphics}
\usepackage{epsfig}
\usepackage{mathptmx}
\usepackage{times}
\usepackage{amsmath}
\usepackage{amssymb}

\usepackage{amsfonts}
\usepackage{pdfx}
\usepackage{booktabs}
\usepackage{tcolorbox}
\usepackage{xspace}
\usepackage{enumitem}
\usepackage{xcolor}
\usepackage{algpseudocode}
\usepackage{wrapfig}
\usepackage{setspace}
\usepackage{soul}
\usepackage{multirow}
\usepackage{inconsolata}
\usepackage{longtable}
\usepackage{tabularx}
\usepackage{array}
\usepackage{marginnote}
\usepackage{cellspace}
\usepackage[vlined,linesnumbered,ruled,resetcount]{algorithm2e}
\usepackage[capitalise, nameinlink]{cleveref}
\usepackage[natbib=true, style=ieee]{biblatex}
\usepackage[subtle,trackingfraction=0.98]{savetrees}
\usepackage[font=footnotesize]{caption}
\usepackage[cal = pxtx]{mathalpha}
\renewcommand{\paragraph}[1]{\noindent \textbf{#1}}

\hypersetup{
    colorlinks=true,
    linkcolor=orange,
    filecolor=magenta,      
    urlcolor=orange,
    citecolor=orange,
}

\newcommand{\system}[0]{RoboCrowd\xspace}

\newcommand{\hiChew}[0]{\texttt{hi-chew}\xspace}
\newcommand{\tootsieRoll}[0]{\texttt{tootsie-roll}\xspace}
\newcommand{\jellyBean}[0]{\texttt{jelly-bean}\xspace}
\newcommand{\hersheyKiss}[0]{\texttt{hershey-kiss}\xspace}
\newcommand{\hiChewBin}[0]{\texttt{hi-chew-bin}\xspace}
\newcommand{\hiChewZiploc}[0]{\texttt{hi-chew-ziploc}\xspace}
\newcommand{\play}[0]{\texttt{play}\xspace}
\newcommand{\tutorial}[0]{\texttt{tutorial}\xspace}
\newcommand{\toolZiploc}[0]{\texttt{tool-ziploc}\xspace}

\newcommand{\hiChewTitle}[0]{Pick up a Hi-Chew\xspace}
\newcommand{\tootsieRollTitle}[0]{Pick up a Tootsie Roll\xspace}
\newcommand{\jellyBeanTitle}[0]{Eject a Jelly Bean from the Candy Dispenser\xspace}
\newcommand{\hersheyKissTitle}[0]{Pick up a Hershey Kiss\xspace}
\newcommand{\hiChewBinTitle}[0]{Pick up a Hi-Chew from the Bin\xspace}
\newcommand{\hiChewZiplocTitle}[0]{Open the Ziploc, Pick up a Hi-Chew, then Close the Ziploc\xspace}

\newcommand{\hiChewDescription}[0]{Move the right arm towards the candy bin. Grasp one Hi-Chew. Drop it in the End Zone. Finally, return to the home position.}
\newcommand{\tootsieRollDescription}[0]{Move the left arm towards the candy bin. Grasp one Tootsie Roll. Drop it in the End Zone. Finally, return to the home position.}
\newcommand{\jellyBeanDescription}[0]{Use the left arm to pull a cup from the cup dispenser. Bring the cup near the lever of the candy dispenser. Use the right arm to align the cup under the lever, then press the lever. Then, use the right arm to pick up the cup and bring it to the End Zone. Finally, return to the home position.}
\newcommand{\hersheyKissDescription}[0]{Move the right arm \textbf{or} the left arm towards the candy bin. Grasp one Hershey Kiss. Drop it in the End Zone. Finally, return to the home position.}
\newcommand{\hiChewBinDescription}[0]{Move the right arm \textbf{or} the left arm towards the candy bin. Grasp one Hi-Chew. Drop it in the End Zone. Finally, return to the home position.}
\newcommand{\hiChewZiplocDescription}[0]{Use the right arm to bring the Ziploc bag to the center of the table. Then, use the left arm to hold the Ziploc while pulling the Ziploc tab with the right arm to open the bag. Then, spread the Ziploc open and pick out a Hi-Chew with the right arm, and bring it to the End Zone. Then, use the right arm to hold the Ziploc while pulling the Ziploc tab closed with the left arm. Finally, use the right arm to place the Ziploc back in the corner of the table, and return the arms to the home position.}

\newcommand{\sceneA}[0]{\texttt{BinScene}\xspace}
\newcommand{\sceneB}[0]{\texttt{Bin+DispenserScene}\xspace}
\newcommand{\sceneC}[0]{\texttt{Bin+ZiplocScene}\xspace}

\newcommand{\shortSceneA}[0]{\texttt{B}\xspace}
\newcommand{\shortSceneB}[0]{\texttt{B+D}\xspace}
\newcommand{\shortSceneC}[0]{\texttt{B+Z}\xspace}

\newcommand{\intuitive}[0]{\texttt{Intuitive}\xspace}
\newcommand{\interesting}[0]{\texttt{Interesting}\xspace}
\newcommand{\wanted}[0]{\texttt{Wanted}\xspace}

\title{{\LARGE \bf \system: Scaling Robot Data Collection through Crowdsourcing} \vspace{-9mm} }

\author{\cr Suvir Mirchandani$^{1}$, David D. Yuan$^{1}$, Kaylee Burns$^{1}$, Md Sazzad Islam$^{1}$, Tony Z. Zhao$^{1}$, Chelsea Finn$^{1}$, Dorsa Sadigh$^{1}$%
\vspace{7mm}
\thanks{$^{1}$Stanford University. Correspondence to {\tt suvir@cs.stanford.edu}.}%
}

\begin{document}

\maketitle

\begin{abstract}
In recent years, imitation learning from large-scale human demonstrations has emerged as a promising paradigm for training robot policies.
However, the burden of collecting large quantities of human demonstrations is significant in terms of collection time and the need for access to expert operators. 
We introduce a new data collection paradigm, RoboCrowd, which distributes the workload by utilizing crowdsourcing principles and incentive design.
RoboCrowd helps enable scalable data collection and facilitates more efficient learning of robot policies.
We build RoboCrowd on top of ALOHA \citep{Zhao2023LearningFB}---a bimanual platform that supports data collection via puppeteering---to explore the design space for crowdsourcing in-person demonstrations in a public environment.
We propose three classes of incentive mechanisms to appeal to users' varying sources of motivation for interacting with the system: material rewards, intrinsic interest, and social comparison.
We instantiate these incentives through tasks that include physical rewards, engaging or challenging manipulations, as well as gamification elements such as a leaderboard.
We conduct a large-scale, two-week field experiment in which the platform is situated in a university caf\'e.
We observe significant engagement with the system---over 200 individuals independently volunteered to provide a total of over 800 interaction episodes.
Our findings validate the proposed incentives as mechanisms for shaping users' data quantity and quality.
Further, we demonstrate that the crowdsourced data can serve as useful pre-training data for policies fine-tuned on expert demonstrations---boosting performance up to 20\% compared to when this data is not available.
These results suggest the potential for RoboCrowd to reduce the burden of robot data collection by carefully implementing crowdsourcing and incentive design principles. Videos are available at \url{https://robocrowd.github.io}.

\end{abstract}

\section{Introduction}
\label{sec:introduction}
With the success of pre-training large models on massive Internet-scale datasets in fields such as natural language processing and computer vision, imitation learning (IL) has become a popular paradigm for training robot policies \citep{Zhao2023LearningFB, rt12022arxiv, rt22023arxiv, octo_2023, kim24openvla}. However, modern IL algorithms continue to have significant data requirements especially as tasks increase in number and variety---on the order of hundreds to thousands of demonstrations. For example, OpenVLA~\citep{kim24openvla} was trained on 970K trajectories from the Open-X Embodiment dataset \citep{open_x_embodiment_rt_x_2023}, much of which was collected by expert human operators over the course of thousands of hours. This underscores the need for scalable methods of collecting robot data.

Prior efforts to scale up real-world data collection range from leveraging videos of human activity~\citep{ma2022vip,nair2022rm} to pooling demonstration data across different institutions~\citep{Khazatsky2024DROIDAL,dasari2019robonet,open_x_embodiment_rt_x_2023}. While the former approach---tapping into internet scale videos---can provide useful visual representations \cite{nair2022r3m,radosavovic2023real,karamcheti2023voltron}, such methods often struggle in tasks beyond pick-and-place without substantial real robot data. On the other hand, pooling datasets across many tasks and embodiments \citep{open_x_embodiment_rt_x_2023,octo_2023} has amortized the cost of real-robot data collection to a degree, but expert operators are still required to collect data especially when new embodiments or tasks are added. Other works focus on how to reduce the time burden on data collectors or guide the collection strategy~\citep{Ross2010ARO,Kelly2018HGDAggerII,gandhi2022eliciting}, but these methods do not address the fundamental problem that demonstrations are still solely collected by researchers or designated operators for the express purpose of training robot policies. This aspect of robot data collection drastically differs from other modalities such as text or images, where large volumes of data are \textit{organically} produced by people in their daily activities and are \emph{readily} available on the web. To explore ways to scale up robot data collection, we ask: \textit{Who} can effectively collect robot data, and \textit{how} might they be incentivized to do so?

\begin{figure}[t]
    \centering
    \includegraphics[width=\linewidth]{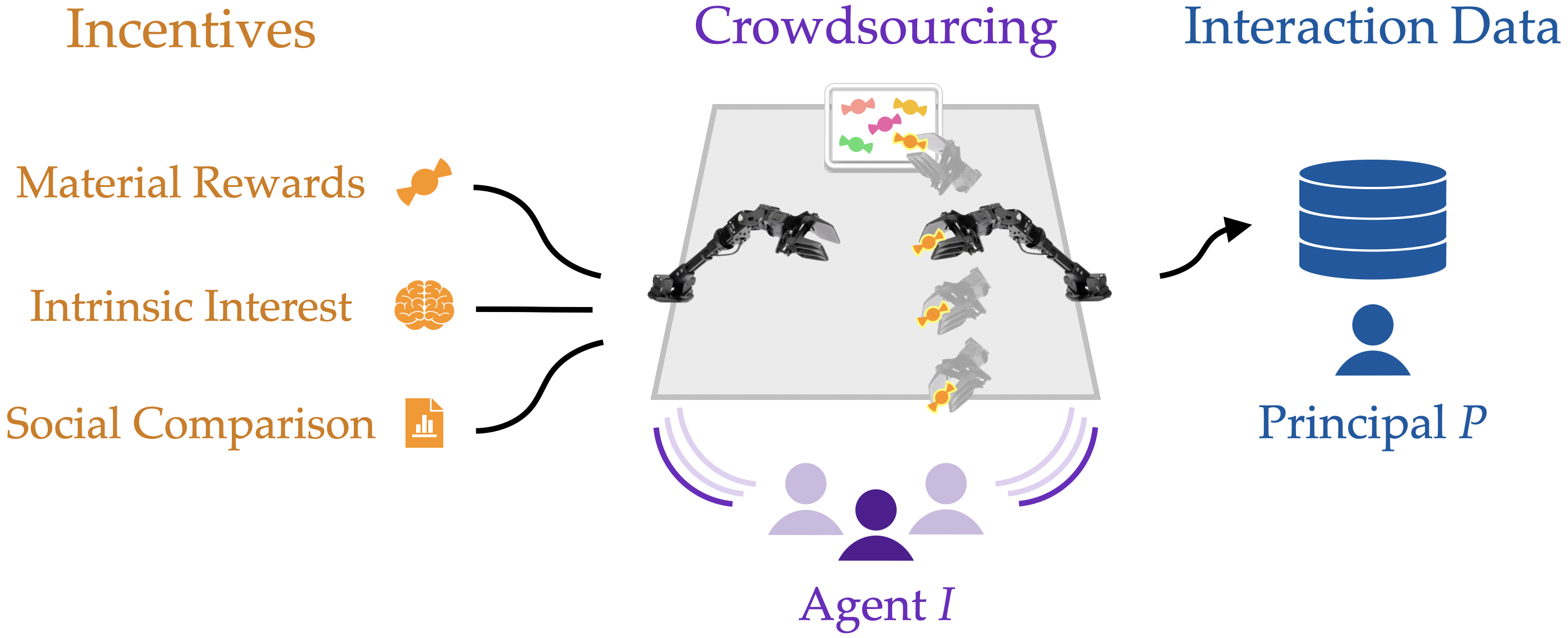}
    \caption{\textbf{Example of incentivizing demonstrations in \system}. The principal $P$ consists of a robot teleoperation setup, a designer, and a scene they have designed. The scene contains tasks that an agent $I$ (a crowd user) can attempt, guided by incentives put in place by the designer. For example, a material reward---e.g., a candy in a bin---can motivate $I$ to produce a successful trajectory for a bin-picking task, which the designer can add to a dataset.}
    \label{fig:front}
\end{figure}

To tackle this problem, we look to a large body of work outside of robotics which studies strategies for incentivizing people in crowdsourced data labeling tasks
~\citep{snow2008cheap, sorokin2008utility, vonAhn2004labeling, bernstein2009collabio, park2019ai, Krishna2016VisualGC}.
The goal of these works is to align the incentives of crowdworkers with researchers' goals of labeling a given dataset---for example, \textit{gamifying} the data labeling process \cite{vonAhn2004labeling} and aggregating data by tapping into the ``wisdom of the crowd.''
Our key idea is to build a system that leverages similar ideas for robot data collection---i.e., \textit{aligning human incentives to provide robot demonstration data}.
However, prior strategies in human-computer interaction are designed for applications that work well with web interfaces, and applying them to robotics introduces several challenges.
First, robot teleoperation traditionally requires access to physical hardware which is not readily available to crowdworkers.
Second, the robot platform must be capable of performing complex tasks---in order to be engaging to users, as well as to collect useful data. At the same time, the system must be intuitive to onboard, since the vast majority of potential data providers have no teleoperation experience. Further, the system must be safe for novice users to operate.

To address these challenges, we propose \system, a framework for incentive design in the context of crowdsourced robot data collection. Our framework centers five key properties: public accessibility, capability, intuitiveness, safety, and gamification.
Diving deeper into the incentive design problem, we incorporate three classes of incentives to appeal to users' varying sources of motivation for interacting with the system. These include \emph{material rewards} (i.e., physical rewards from tasks), \emph{intrinsic interest} (i.e., motivation from engaging tasks), and \emph{social comparison} (i.e., comparison to other users).
To instantiate the framework, we build upon ALOHA \cite{Zhao2023LearningFB}---a bimanual platform for robot teleoperation---to satisfy our need for capable hardware, and situate the system in a public space to enable access to general users. We design hardware enhancements and a user interface to make teleoperation intuitive and safe for non-experts.
\cref{fig:front} illustrates an example of how incentives might shape user interactions with the system into useful data: a scene contains a bin of candies, and when a user successfully acquires a candy via teleoperation, they simultaneously contribute a trajectory to a bin-picking dataset.

We deploy the system in a field experiment in which the robot is situated near a university café, where users participate in a self-guided, gamified data collection experience. We observe significant engagement with the system---over 200 individuals independently volunteered to provide a total of over 800 interaction episodes. We compile the crowdsourced interactions into a dataset and annotate each trajectory with quality scores and task labels. We additionally validate material rewards, intrinsic interest, and social comparison as incentive types for shaping user interactions with the robot---observing up to 2$\times$ the amount of data collection time spent on tasks that have preferred physical rewards (when controlling for task type) and up to 4$\times$ the amount of data collection time spent on tasks that are more engaging (when controlling for physical reward). We additionally observe a positive correlation between users' response to a leaderboard (social comparison mechanism) and their data quality and quantity. Finally, we analyze the usefulness of the crowdsourced data for training policies. We demonstrate that the crowdsourced data can serve as useful pre-training data when fine-tuning on expert demonstrations, boosting policy performance up to 20\% compared to expert-only policies.
To our knowledge, \system is the first system to crowdsource in-person demonstrations for imitation learning directly from a public audience---with the potential to enable a new avenue for scalable robot data collection.

\section{Related Work}
\label{sec:related_work}

In this section, we provide an overview of prior work in crowdsourcing data collection and labeling in robotics and other fields. \looseness=-1

\paragraph{Crowdsourcing Non-robot Data.}
Crowdsourcing is a well-studied technique in human-computer interaction, often used for collecting data labels from a large set of users, with a variety of applications from computer vision to natural language processing~\citep{Krishna2016VisualGC, 10.1371/journal.pcbi.1006337, ørting2019survey, Cenggoro_2018, 10.1007/978-3-030-44429-7_11, Shmueli_2021, nangia2021ingredients, mishra2022crosstask, lim-etal-2020-annotating, 10.1145/2470654.2466265}.
While many works utilize platforms such as Amazon Mechanical Turk \cite{berlin2012shaping} and Prolific \cite{palan2018prolific} to pay crowdworkers for data labels, other works consider how to \textit{incentivize} crowdworkers via other incentives beyond direct payment to gather data \cite{von2008recaptcha, bernstein2009collabio, law2011human, 7384520, huynh2021survey}.
For example, Games-with-a-Purpose (GWAPs) \cite{vonAhn2004labeling} utilize gamification---the use of game-like elements in non-game contexts \cite{deterding2011game}---to guide users to give higher quality data labels. In \cite{vonAhn2004labeling}, two players try to agree on words to describe pictures without otherwise communicating---resulting in quality image label data.
Our work aims to investigate how incentive design can be adapted and applied to robot data collection---specifically, in the form of demonstrations collected via teleoperation.

\paragraph{Distributing Robot Data Collection.}
Crowdsourcing has been an attractive approach for collecting data in robotics in recent years. Prior works have attempted to crowdsource robot data via remote teleoperation in simulation or via web interfaces. RoboTurk \cite{Mandlekar2018ROBOTURKAC, mandlekar2019scaling} develops a smartphone interface to allow crowdworkers on Mechanical Turk to collect demonstrations remotely, and shows the potential of using crowdsourced data to aid policy learning. 
Although this method alleviates the need for crowdworkers to physically interact with robot hardware, the ability to perform precise tasks can be limited (due to issues such as lack of depth perception, occlusion, etc.). It also presents challenges in recovering from failure states in real-world scenarios.
Other works have utilized crowdsourcing to guide exploration in real-world reinforcement learning~\citep{Balsells2023AutonomousRR,torne2023breadcrumbs} or to collect interaction data through high-level abstractions \cite{chung2014accelerating, van2021exploring, chernova2011crowdsourcing}, but are again limited in the range of tasks that can be collected because they do not focus on low-level trajectory demonstrations. 

Several works have developed new interfaces to make robot demonstration collection more distributed.
Recent works \citep{song2020grasping, wang2024dexcap, Chi2024UniversalMI} design new hardware interfaces---e.g., sensorized hand-held grippers or portable motion capture systems---to allow for demonstration collection in the real-world without needing access to a physical robot.
However, crowdsourcing data with these interfaces is not immediately possible since it still requires data collectors to have access to this custom hardware.
\citep{wang2024eve} presents an augmented reality tablet interface to collect robot data from everyday users, though it does not immediately extend to bimanual or dynamic tasks.
In this work, rather than introducing a new teleoperation interface, we leverage an existing interface (puppeteering via ALOHA~\citep{Zhao2023LearningFB}, which enables precise bimanual manipulation at a low-cost) and choose to situate it directly in a public space to make it accessible to data collectors. To make scaling up data collection possible, we design the system so it can be used by non-experts. %

\section{Preliminaries}
\label{sec:preliminaries}
In this section, we provide an overview of the problem of designing incentives for crowdsourcing data collection and the problem of imitation learning from collected demonstrations.

\paragraph{Crowdsourcing via Incentive Design.}
Crowdsourcing systems can be modeled as repeated principal-agent interactions. We adapt the notation from \cite{ratliff2019perspective}. A principal
$P$ desires a pool of tasks $\mathbb{T}$ to be completed by an agent (or set of agents) $I$ with maximum quality at minimum cost.
$P$ and $I$ each have utility functions, denoted as $J_P: A_I \times A_P \to \mathbb{R}$ and $J_I: A_I \times A_P \to \mathbb{R}$, where $A_I$ and $A_P$ are the action spaces of the agent and principal respectively. An \textit{incentive} $\gamma: A_I \to A_P$ maps between agent actions and principal actions. $P$ aims to design $\gamma$ to shape $I$'s actions in a way that maximizes $J_P$, noting that for any given $\gamma$, agents have utility $J_I(a_I, \gamma(a_I))$.

In the context of crowdsourcing robot data, $P$ abstractly represents the data collection platform and its designer, and $I$ represents a user in the presence of the platform. 
For the principal $P$, $A_P$ encapsulates all actions that the robot can take and how the scene changes in response to robot actions.
$J_P$ corresponds to how \textit{useful} the data collected by $I$ is to $P$ towards constructing a crowdsourced dataset $\mathcal{D}$. %
In this work, we quantify this notion in several ways (number of demonstrations, length of demonstrations, human-labeled quality scores, and downstream policy learning performance).
For the agent $I$, $A_I$ defines $I$'s possible actions, such as the task choice and teleoperation actions. $J_I$ is the utility derived by the agent from intrinsic and extrinsic factors when interacting with the robot. $J_I$ is multifaceted and can vary widely for each $I$. Towards our goal of crowdsourcing data, we explore different facets of utility, such as the utility derived from receiving a physical reward as an outcome for completing a task, intrinsic interest in the task itself, and motivation driven by social comparison.

An incentive $\gamma$ is a mapping from $A_I$ to $A_P$. This mapping is induced by a set of decisions that $P$ makes in developing the robot's \textit{scene context}, such as the tasks available.
To illustrate, consider a scene context consisting of a robot and a bin of physical rewards (e.g., candies), as in \cref{fig:front}. In response to an agent action $a_I$ (e.g., teleoperating the robot to handover a reward), the principal takes an action $a_P$ (the robot moving as directed) which results in the agent receiving the reward. The reward is factored into the agent's utility $J_I(a_I, \gamma(a_I))$.

When the existence of an incentive $\gamma$ affects $I$'s actions such that both $J_I$ and $J_P$ increase, the incentive is \textit{aligned} between $P$ and $I$. For example, if $I$ prefers to receive a candy, and teleoperates the robot in order to acquire a candy, $J_I$ increases by the value of one candy and $J_P$ increases in that there is one more trajectory to include in the crowdsourced dataset $\mathcal{D}$. We instantiate incentives of different classes, and illustrate that they are effective mechanisms for shaping the quality and quantity of behaviors in $\mathcal{D}$. We next describe how policies can be learned from $\mathcal{D}$ via imitation learning.

\paragraph{Imitation Learning.}
Imitation learning (IL) aims to learn a policy $\pi_\theta$ parameterized by $\theta$ from a dataset $\mathcal{D}$ composed of expert demonstrations. Each demonstration $\xi \in \mathcal{D}$ is a sequence of observation-action transitions $\{(o_0, a_0), \ldots, (o_T, a_T)\}$. Most commonly, IL is instantiated as behavior cloning, which trains $\pi_\theta$ to minimize the negative log-likelihood of data, $\mathcal{L}(\theta) = -\mathbb{E}_{(o,a)\sim \mathcal{D}}[\log \pi_\theta (a | o)]$. Since human-collected demonstrations may be diverse in practice, algorithms such as Action Chunking with Transformers (ACT) \citep{Zhao2023LearningFB} are designed to model different modes of behavior. We provide an overview of ACT in \cref{appx:act}.
The success of this training paradigm hinges on the quality and quantity of trajectories in $\mathcal{D}$. We frame the creation of $\mathcal{D}$  through the lens of crowdsourcing and incentive design.

\section{\system}
\label{sec:system}

\begin{figure*}
    \centering
    \includegraphics[width=0.9\linewidth]{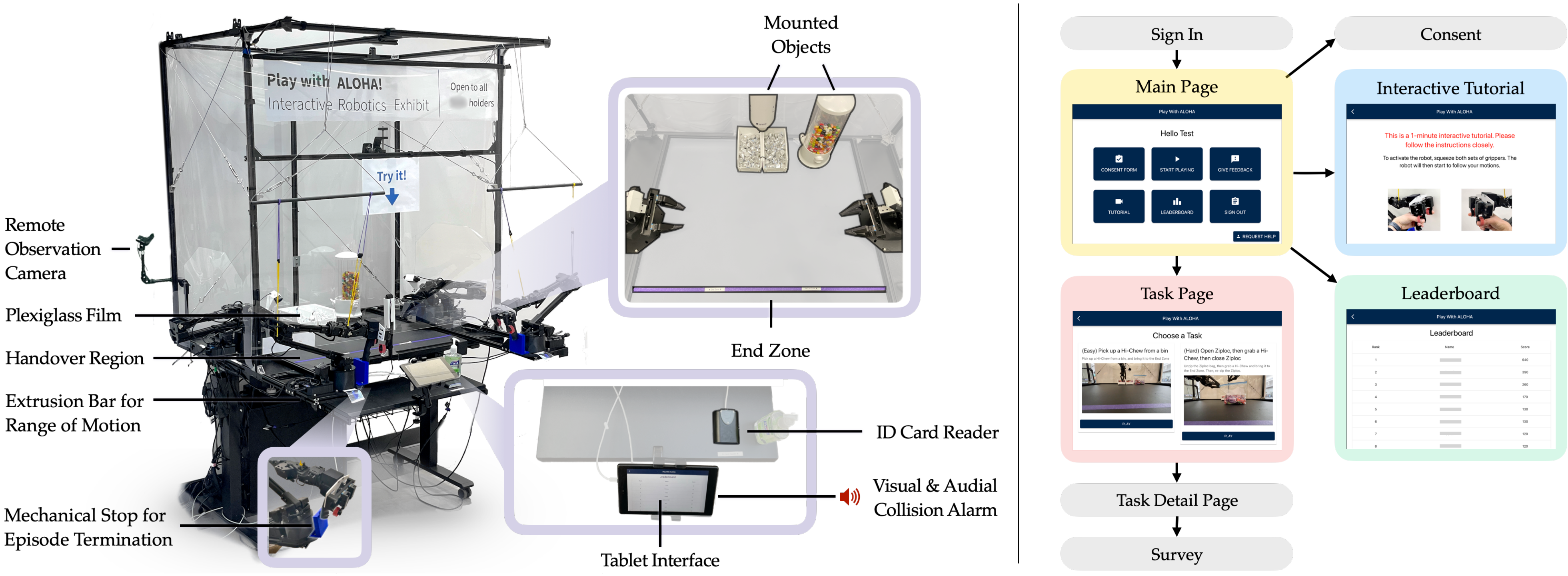}
    \caption{\textbf{System Overview}. (\textit{Left}) RoboCrowd uses the ALOHA robot \cite{Zhao2023LearningFB}, a bimanual teleoperation platform wherein users control 2 ViperX follower arms by puppeteering via 2 WidowX leader arms. Users can perform tasks in scenes put in place by the scene designer; tasks may include physical rewards that the user can bring to the End Zone and access via the Handover Region. (\textit{Right}) Users are guided by a GUI on a tablet. Functionalities include an Interactive Tutorial to get acquainted with \system, a Task Page to select among tasks, and a Leaderboard where users can compare their scores. For additional details, please see \cref{appx:software}.}
    \label{fig:system_overview}
    \label{fig:interface}
    \vspace{-2mm}
\end{figure*}

In this work, we apply incentive design to the collection of robot demonstrations for imitation learning, and develop a system to collect robot demonstrations directly from the public. We propose three different incentive mechanisms to appeal to users' varying utility functions, and illustrate that incentives can impact the quantity and quality of data collected. Finally, we demonstrate the usefulness of the data for policy learning.

\paragraph{Enabling In-Person Crowdsourced Teleoperation.} While a sizeable body of work has studied crowdsourcing and incentive design in the context of data labeling, applying these ideas to robot demonstration collection introduces numerous challenges. First, members of the public lack direct access to robots.
Additionally, implementing incentives appropriate for real-world robot demonstrations---e.g., physical rewards and intrinsically interesting tasks---requires hardware that is capable of versatile tasks.
Finally, the majority of potential users lack experience teleoperating robots, so the system must be easy and safe for users to use.
Given these challenges, we establish a set of desired properties for our system to enable crowdsourcing robot data as well as to effectively implement the incentive mechanisms described subsequently. \looseness=-1
\begin{itemize}[leftmargin=18px]
    \item [P1] \textit{Publicly Accessible.} The system should be open to members of the public, including non-roboticists.
    \item [P2] \textit{Capable}. The hardware should be capable of performing complex manipulation skills.
    \item [P3] \textit{Intuitive}. The system should be intuitive to novice users with a self-guided onboarding process.
    \item [P4] \textit{Safe}. The system should be safe for novice users to operate.
    \item [P5] \textit{Gamified.} The system should permit gamified elements---e.g., the ability to track individual users and the ability to provide physical rewards.
\end{itemize}

\paragraph{Designing Incentive Mechanisms.}
Given a system that crowdworkers can interact with, we design incentive mechanisms to shape these interactions into useful data. We expect that crowdworkers may vary in their utility functions $J_I$. Some may be motivated by extrinsic rewards whereas others may be intrinsically interested in challenging tasks. Still others---e.g., people who are more competitive---may be motivated by social comparison \cite{kruglanski1990classic}.
We therefore design for three incentive mechanisms:
\begin{itemize}[leftmargin=18px]
    \item [M1] \textit{Material Rewards}. Designing a scene with material rewards means that for some agent actions $a_I$, $P$ performs actions $\gamma(a_I) \in A_P$ such that $I$ receives a physical object. For example, $I$ teleoperating a bin-picking task results in $P$ performing an action which delivers the reward to $I$. 
    \item [M2] \textit{Intrinsic Interest}. Designing a scene for intrinsic interest expands $A_P$ to include engaging or challenging tasks. For example, as the result of certain agent teleoperation actions $a_I$, $P$ may perform fine-grained object manipulation $\gamma(a_I)$.
    \item [M3] \textit{Social Comparison}. Designing a scene to enable social comparison involves a mechanism where agents can compare themselves. For example, the action $a_I$ of teleoperating a successful trajectory can result in a principal action $\gamma(a_I)$ which awards points to $I$ and increases their leaderboard position.
\end{itemize}

The rest of this section explains how we meet these desiderata through our hardware and software design.

\subsection{Hardware Design}

We select ALOHA \cite{Zhao2023LearningFB}, a system for bimanual teleoperation, as the base platform for our system.
ALOHA consists of two ``follower'' arms (ViperX) that are controlled via puppeteering with two ``leader'' arms (WidowX). We choose to use the ALOHA platform due to its low-cost, repairability, as well as its ability for collecting data for a wide task range. \cref{fig:system_overview} illustrates a set of enhancements to outfit ALOHA for public use to achieve our desired properties and enable crowdsourcing. First, we implement mechanisms for user and robot safety (P4): (a) collision avoidance to prevent self-collisions, achieved via a parallel MuJoCo \cite{todorov2012mujoco} simulator, as well as a visual-audial alarm when the robot is near collision; (b) plexiglass and vinyl film to cover all sides of the ALOHA workcell to enclose the puppet arms; (c) extended extrusion bars on the leader arms to increase the distance between users and leader arms; (d) mounting of scene props (such as bins and dispensers) to mitigate scene damage; and (e) a remote observation camera for the scene designer to periodically monitor the scene. We also include enhancements to increase the intuitiveness of the platform for members of the public (P3): (a) a tablet interface, described in the next section; (b) a mechanical stop for users to automatically terminate episodes by resting the puppet arms. To enable a gamified setup (P5), we utilize (a) an ID card reader to authenticate and track users and (b) demarcate an ``End Zone'' within scenes, where a user a can place physical rewards and access them via a handover region at the bottom of the plexiglass casing. Given its ability to perform versatile tasks, ALOHA satisfies our capability goal (P2). We physically situate it in a public environment (\cref{sec:experimental_setup}) to make it accessible to crowd users (P1).

\subsection{Software Design}

To make operating the robot intuitive (P2) for members of the public, we implement a tablet application to complement the hardware platform and guide users through the operation process (\cref{fig:interface}; right). The interface additionally features a variety of elements of gamification (P5) that we highlight below. 

\paragraph{Onboarding.} We develop an onboarding process for new users to sign-in and receive a tutorial to familiarize themselves with the platform. In pilot studies (\cref{appx:pilots}) where users were asked to use the system but were not given further verbal instructions (to mimic organic encounters that crowd users might have), users reported a desire for ``instant gratification'' and wished to begin to use the robot as soon as possible rather than watching a video or reading instructions. Thus, we design our onboarding process to be efficient and interactive: users begin by tapping their university ID card on a card reader, which directs them to a Sign In page to create a \textit{user profile}. Users are then directed to complete a consent form and an interactive tutorial to learn how to puppeteer the robot (\cref{fig:interface}; right). The tutorial contains four steps and takes less than one minute to complete. We detail the stages of the interactive tutorial in \cref{appx:software}.

\paragraph{Performing Tasks.} After completing the tutorial, users can choose to enter a \textit{Task Page} where they see videos of different tasks they can complete in the scene (\cref{fig:interface}; right). These tasks can be presented in various ways; for example, marked with \textit{levels of difficulty} (e.g., easy versus hard). In service of P5, we use gamified verbiage and elements throughout the interface (e.g. a \textit{Start Playing} button, and a \textit{countdown timer} on performing tasks). Specifically for M3, we implement a point system where users receive points for completing tasks, which are tallied and visible on a \textit{Leaderboard Page}, where users can see how their scores rank compared to other users (\cref{fig:interface}; right). For details on the software architecture, please see \cref{appx:software}.

\begin{figure}
    \centering
    \includegraphics[width=\linewidth]{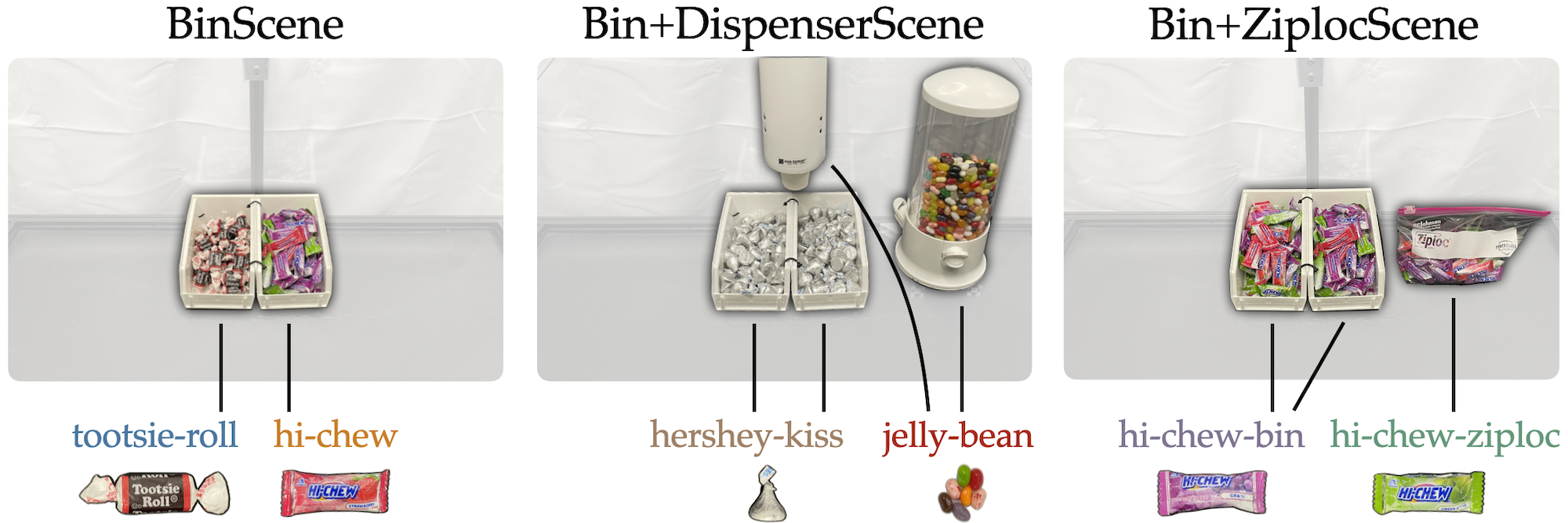}
    \caption{\textbf{Scene Setup}. Illustration of \sceneA, \sceneB, and \sceneC, and the objects relevant to our 6 tasks (\hiChew, \tootsieRoll, \hersheyKiss, \jellyBean, \hiChewBin, \hiChewZiploc).}
    \label{fig:scenes}
\end{figure}

\section{Experimental Setup}
\label{sec:experimental_setup}
We utilize RoboCrowd to collect a crowdsourced dataset over a two-week period in a public university caf\'e. We instantiate M1--M3 to appeal to users' varying utility functions $J_I$, and design scenes in order to verify if these mechanisms can shape demonstration quantity and quality.
This section details our experimental setup. We analyze the data and discuss the results in~\cref{sec:results}.

\paragraph{Scene Design.}
On each day of crowdsourcing, two of six tasks are made available to users, with different pairs corresponding to different scenes (\cref{fig:scenes}).
\sceneA contains bins with two types of candies for single arm bin-picking tasks (\hiChew and \tootsieRoll). \sceneB contains the same bins with a single type of candy (\hersheyKiss), as well as a cup dispenser and a jelly bean dispenser (\jellyBean). \sceneC contains the same bins with a single candy type (\hiChewBin) as well as a closed Ziploc bag full of candies (\hiChewZiploc).

\paragraph{Incentive Types.}
We design tasks within the aforementioned scenes in order to study incentives M1--M3, as we describe below.

[M1] \textit{Material Rewards.}
We hypothesize that direct material rewards can influence which tasks users perform with the system.
We design a simple scene context to test this (\sceneA). There are two bins on the table, one containing Hi-Chews and the other containing Tootsie Rolls (\cref{fig:scenes}).
There are two bin-picking tasks available to the user on the Task Page: ``pick up Hi-Chew'' (\hiChew) and ``pick up Tootsie Roll'' (\tootsieRoll).
We hypothesize that users who engage with the robot will more often choose to interact with the Hi-Chew (which, in an offline survey, we find is more desired than the Tootsie Roll; see \cref{appx:dataset_analysis} for more discussion on our choice of tasks). 
This incentive mechanism is an example of \textit{extrinsic motivation}.

[M2] \textit{Intrinsic Interest.}
Users may also be \textit{intrinsically} motivated in how they choose to interact with the system. We hypothesize that users prefer to spend time on tasks that are more interesting and challenging.
Thus, we design \sceneC to contain a bin with Hi-Chews as well as a closed Ziploc bag with Hi-Chews inside. This scene features two available tasks: ``pick up Hi-Chew from bin'' and ``open Ziploc, pick up Hi-Chew, close Ziploc.''
With the same extrinsic reward, the latter task is \textit{significantly} more challenging, yet may be more intrinsically interesting to users.
We test this effect in \sceneB as well, which contains a bin with Hershey Kisses as well as a cup dispenser and a dispenser  containing Jelly Beans. The tasks available in this scene are ``pick up Hershey Kiss from bin'' (\hersheyKiss) and ``take cup from dispenser and eject Jelly Bean into the cup'' (\jellyBean). The latter task is again significantly more challenging; but note that it does not provide greater extrinsic reward according to our offline survey (see \cref{appx:dataset_analysis}).

[M3] \textit{Social Comparison.}
Users may vary in how they respond to gamification mechanisms for social comparison in the interface. To test the idea that gamified elements can shape the way certain users collect data, we include a leaderboard that tallies the number of ``points'' users achieve by completing tasks (\cref{fig:interface}).
Using quantitative measurements to compare players, including via leaderboards, is a common method for provoking competition \cite{medler2011analytics}.
We hypothesize that users who choose to look at the leaderboard may give a higher quantity of data, stemming from social comparison as an incentive mechanism.

\paragraph{Data Annotation Pipeline.}
User interaction data are a mixture of task-relevant data, tutorial interactions, and ``play'' data.
We manually annotate all interactions by whether the user was engaging in free play or task-relevant behavior, as well as quality scores on a scale of 0 (play data) to 3 (highest quality task data). We define quality labels based on multiple factors as detailed in \cref{appx:software}, including how smooth the user's motions are, whether there is retrying behavior, etc.
Importantly, each interaction episode may include data relevant to different tasks and of various qualities, so we annotate these labels at \textit{every transition} per trajectory. 

\paragraph{Metrics.}
We analyze the crowdsourced data using several metrics:

\begin{itemize}[leftmargin=*, ]
\item \textit{Quantity.} Our primary metric measures the number of timesteps a user spends performing a task.

\item \textit{Quality}. We utilize our data quality annotations, and additionally explore other data quality measures in \cref{appx:dataset_analysis}.

\item \textit{Usefulness for Policy Learning.} We study the utility of the crowdsourced data for policy learning via co-training and fine-tuning with expert demonstrations.

\item \textit{Self-reported Likert Ratings.} We survey users for self-ratings of intuitiveness, enjoyment, and how well the robot completed the task in the way they desired, and report results in \cref{appx:dataset_analysis}.

\end{itemize}

\section{Results}
\label{sec:results}
We analyze the composition of the dataset, effects of incentive mechanisms, and usefulness of the data for policy learning.

\subsection{Usage Overview and Dataset Composition}

We observe significant engagement with RoboCrowd over the two-week collection period: there were $N = $ 231 unique users in total.
On most days, more than two-thirds of these were new users that had not used the system on prior days. There were a total of 817 interaction episodes distributed throughout the period.

\begin{figure}
    \centering
    \vspace{2mm}
    \includegraphics[width=\linewidth]{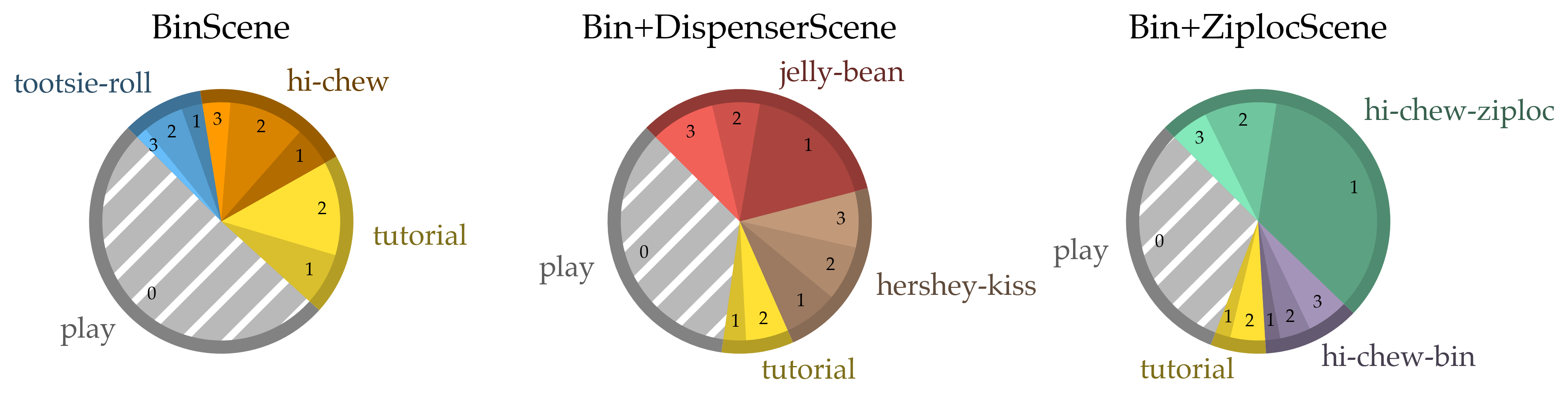}
    \caption{\textbf{Dataset composition by number of time steps for each of our three scenes}. Different hues indicate different tasks. Tasks receive quality scores from 1 to 3 (higher is better) which are also indicated by brighter shades. Tutorial data receives a score of 1 or 2. Play data always receives a score of 0.}
    \label{fig:composition}
\end{figure}

Our dataset is composed of 3 scenes (\cref{fig:scenes}). We collect 129 interaction episodes in \sceneA (\shortSceneA; Day 1), 381 in \sceneB (\shortSceneB; Days 2--5), and 307 in \sceneC (\shortSceneC; Days 6--11). In aggregate, users spent 54.2\% of interaction time performing the preset tasks in the scene, 9.6\% on the interactive tutorial, and 36.1\% on free-play. While we focus our learning experiments on task-relevant data in \cref{sec:policy-learning}, this play data could be fruitful for training multitask policies in the future.
In~\cref{fig:composition}, we show the distribution of tasks and qualities over timesteps for each scene. Qualities are determined on a scale from 1--3 for task-relevant data and a scale of 1--2 for tutorial data based on the smoothness of the user's motion and whether there is retrying behavior or extraneous movements. We detail the quality annotation rules in \cref{appx:software}, and illustrate sample trajectories in each scene in \cref{appx:task_details}.

\subsection{Effects of Incentives on Data Quantity and Quantity}

\textit{Material Rewards.}
While \sceneA contains two bin-picking tasks with nearly identical difficulty, users in aggregate spend 2$\times$ as many timesteps performing \hiChew compared to \tootsieRoll. This suggests that users devote more interaction time to tasks where the direct material incentive is more preferred. We also find that users spend a significant amount of time (50.7\%) in free-play with the system in \sceneA, engaging in behaviors such as trying out more challenging tasks (e.g., attempting to unwrap the candies; see \cref{appx:dataset}). Thus, while material incentives can influence user demonstrations (e.g., higher material incentives can lead to more data), drivers of intrinsic motivation such as the difficulty of the task also play a role, as we discuss next.

\textit{Intrinsic Motivation.}
Interestingly, in \sceneB, which contains a harder bin-picking task than in Scene A (\hersheyKiss) and a challenging long-horizon candy dispensing task (\jellyBean), users spend only 35.3\% of the time in free-play. Additionally, despite the fact that users do not generally prefer Jelly Beans over Hershey Kisses as a material reward, they still spend more (1.5$\times$) time performing the \jellyBean task. This suggests that intrinsic interest can influence users to allocate more time doing harder task compared to easier ones, or engaging in free-play. To probe whether this intrinsic motivation effect is present even when controlling for the material reward, we consider \sceneC.
 Here, the incentive is contained within a closed Ziploc bag which must be opened. The same incentive is available in the bin to be picked.
 Users spend $4.18\times$ as many timesteps on \hiChewZiploc compared to \hiChewBin, again suggesting that intrinsic motivation influences which tasks users perform in the scene.

\begin{wrapfigure}[9]{R}{0.47\linewidth}
    \centering
    \vspace{-15pt}
    \includegraphics[width=\linewidth]{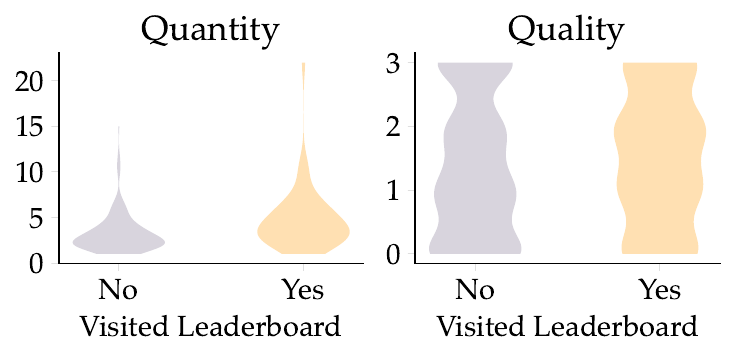}
    \caption{\textbf{Quantity and quality by leaderboard use}. Violin plot showing the distribution of quantity and quality of demonstrations for users who did and did not visit the leaderboard.}
    \label{fig:leaderboard}
\end{wrapfigure}
\textit{Social Comparison.}
To examine how different people respond to explicit comparison mechanisms in the system, we record which users visit the Leaderboard, and conduct a Mann-Whitney U-test to compare the quantity and quality of demonstrations provided by Leaderboard visitors compared to other users. \cref{fig:leaderboard} illustrates the distribution of quality (number of interactions) and quality (mean quality score) conditioned on Leaderboard visitation. We find that visitors of the Leaderboard provide significantly more demonstrations ($p<0.001$) that are higher quality on average ($p<0.05$). \looseness=-1

\subsection{Policy Learning with the Crowdsourced Data}
\label{sec:policy-learning}

\begin{table}
    \centering
    \vspace{2mm}
    {
    \scriptsize
    \begin{tabular}{cccccc} \toprule
        Task & Scene & \# Exp. & Expert & Co-train & Fine-tune \\ \midrule
        \hiChew & \shortSceneA & 30 & 37.5\% & 27.5\% & \textbf{42.5\%} \\
        \tootsieRoll & \shortSceneA & 30 & \textbf{42.5\% }& 25\% & 40\%\\
        \hersheyKiss & \shortSceneB & 60 & 20\% & 32.5\% & \textbf{35\%} \\ 
        \hiChewBin & \shortSceneB & 80 & 20\% & 12.5\% & \textbf{40\%} \\ 
        \jellyBean & \shortSceneC & 100 & \textbf{{48.9 $\pm$ 18.6}} & {8.9 $\pm$ 10.1} & {19.7 $\pm$ 29.7} \\
        \hiChewZiploc & \shortSceneC & 100 & {5.4 $\pm$ 12.2} & {17.1 $\pm$ 15.8} & \textbf{{22.1 $\pm$ 14.3}} \\ \bottomrule
    \end{tabular}
    }
    \vspace{2mm}
    \caption{\textbf{Policy Performance.} Performance of policies trained on expert demonstrations (\# Exp.), co-trained on crowd data, and pre-trained on expert+crowd data then fine-tuned on expert data. We conduct 40 trials for each cell. For the long-horizon tasks (\jellyBean, \hiChewZiploc), we provide a normalized return (out of 100) rather than success rate (see \cref{appx:policy_learning} for details).}
    \label{tab:policy_learning}
\end{table}

Finally, we study how useful the crowdsourced data is for policy learning.
To complement this dataset, we collect a set of expert demonstrations: 30 demonstrations for each of \hiChew and \tootsieRoll, 60 for \hersheyKiss, 80 for \hiChewBin, and 100 for each of \jellyBean and \hiChewZiploc. \looseness=-1

In \cref{tab:policy_learning}, we compare different methods of mixing crowdsourced data and expert data on our six tasks. All policies use ACT \cite{Zhao2023LearningFB} with default hyperparameters.
Training exclusively with the expert data on each task constitutes the \textit{Expert} setting. \textit{Co-train} refers to na\"ively mixing data from a crowdsourced task (i.e., task-relevant data of any quality) with the expert data. We also compare to \textit{Fine-tune}, which trains in two stages: first co-training on the crowd data and expert data and then fine-tuning on expert data only; for fair comparison, note that \textit{Fine-tune} is trained for fewer total steps (150K) than both \textit{Expert} and \textit{Co-train} (200K). Crowdsourced data provides performance improvements in multiple cases, but the specific effects vary by task. For example, crowdsourced data can involve low-quality behaviors (e.g., regrasping behavior or grasping multiple items at a time), which may cause the \textit{Co-train} to perform worse than \textit{Expert}, but still provide a useful initialization for \textit{Fine-tune}; one exception is \jellyBean, for which the crowdsourced data is especially diverse. We provide additional qualitative analysis of the trained policies in \cref{appx:evaluation-details}. \looseness=-1

\begin{wraptable}[9]{r}{0.5\linewidth}
    \centering
    \vspace{-3mm}
    {
    \scriptsize
    \begin{tabular}{ccc} \toprule
        Subtask & Expert & Fine-tune \\ \midrule
        Grasp Corner & 85\% & \textbf{95\%} \\
        Pinch Tab & 70\% & \textbf{85\%} \\
        Slide Open & 60\% & \textbf{80\%} \\ \bottomrule
    \end{tabular}
    }
    \vspace{2mm}
    \caption{Staged success rate for a policy pre-trained on \hiChewZiploc crowd data and fine-tuned on 50 expert demos of \toolZiploc compared to expert-only \toolZiploc policy.}
    \label{tab:downstream_finetuning}
\end{wraptable}
We also demonstrate that the crowdsourced data can benefit downstream tasks. In \cref{tab:downstream_finetuning}, we train an expert-only ACT policy (50 demos) to convergence on a new task, \toolZiploc, which requires unzipping a Ziploc containing tools. The unzipping skill is shared with \hiChewZiploc. We compare this expert-only policy to a policy trained with two stages (pre-training on crowdsourced \hiChewZiploc data and then fine-tuning on the expert \toolZiploc data). This outperforms the expert-only policy by 20\%, suggesting that crowdsourced data can be beneficial in downstream tasks with shared manipulation skills. \looseness=-1

\section{Discussion and Limitations}
\label{sec:discussion}
In this work, we propose a new paradigm for robot data collection via crowdsourcing and incentive design.
We focus on three incentive types---material rewards, intrinsic motivation, and social comparison---but there are further avenues to explore within these categories as well (e.g., how physical rewards differ from monetary incentives).
Crafting data collection schemes where people are motivated by external rewards, fun, interest, or competition is a general principle, and a rich area for future work would be to scale up our findings on incentive design in robot data collection to new tasks. For example, appealing to extrinsic motivation and social comparison could help craft a data collection scheme for a task such as packing groceries---where users are motivated by spaced rewards (getting to keep every $N$ bags) or social comparison (getting points for more efficient packing). A variety of other incentive types (e.g., task novelty, collective effort, robot's ability to learn from the data, etc.) could be applied to new settings as well.
While crowdsourcing has the benefit of reducing data collection effort of individual researchers, it also presents challenges of data quality control and heterogeneity, especially as tasks increase in complexity.
We hope that our dataset---collected from over 200 users with manual fine-grained quality annotations---can be helpful to future works seeking to understand the style and diversity of different human operators, and what the most effective ways are to leverage crowdsourced data during downstream policy learning.

\section*{Acknowledgments}

We are grateful for support from Toyota Research Institute, ONR Award N000142212293, NSF Awards 2132847,  2218760, and 2006388, DARPA Award W911NF2210214, DARPA TIAMAT project, and the Stanford Human-Centered AI Institute Hoffman-Yee Grant. We thank members of the Stanford ILIAD and IRIS labs for useful discussions and feedback.

\printbibliography

\clearpage

\appendix
\appendices
\captionsetup{aboveskip=10pt,belowskip=20pt}

\section*{Overview}

In the appendices below, we provide additional details on the implementation of \system, our experiments, and our crowdsourced dataset. We provide a brief overview of each appendix below. %

\medskip

\noindent \cref{appx:task_details} -- \textit{Task Details}

\begin{minipage}{.95\linewidth}
    \medskip
    We give descriptions of each of our 6 tasks, as well as renderings and images depicting sample expert demonstrations for each task.
    \medskip
\end{minipage}

\noindent \cref{appx:dataset} -- \textit{Dataset Examples}

\begin{minipage}{.95\linewidth}
    \medskip
    We provide sample trajectories from our collected dataset including their task and quality annotations, to qualitatively illustrate the diversity of the behaviors in the dataset.
    \medskip
\end{minipage}

\noindent \cref{appx:dataset_analysis} -- \textit{Additional Dataset Analysis}

\begin{minipage}{.95\linewidth}
    \medskip
    We provide further data analysis, including an offline user study to justify our scene choices, additional data quality analysis, and results on users' self-reported Likert ratings of their interactions with the system.
    \medskip
\end{minipage}

\noindent \cref{appx:policy_learning} -- \textit{Additional Details on Policy Learning Experiments}

\begin{minipage}{.95\linewidth}
    \medskip
    We provide additional details on the training and evaluation procedures for our policy learning experiments, as well as further qualitative analysis of the results.
    \medskip
\end{minipage}

\noindent \cref{appx:software} -- \textit{Additional Details on Software Implementation and Data Annotation}

\begin{minipage}{.95\linewidth}
    \medskip
    We provide further details on the graphical user interface, interactive tutorial, software implementation, and data annotation pipeline.
    \medskip
\end{minipage}

\noindent \cref{appx:pilots} -- \textit{Additional Details on Pilot Studies and System Development}

\begin{minipage}{.95\linewidth}
    \medskip
    We provide more details on how we designed and refined the system through pilot studies.
    \medskip
\end{minipage}

\noindent \cref{appx:act} -- \textit{Overview of Action Chunking with Transfomers (ACT) \cite{Zhao2023LearningFB}}

\begin{minipage}{.95\linewidth}
    \medskip
    We provide additional background on the Action Chunking with Transfomers (ACT) algorithm.
    \medskip
\end{minipage}

\label{appx:overview}

\section{Task Details}
\label{appx:task_details}
In \cref{tab:hi-chew,tab:tootsie-roll,tab:jelly-bean,tab:hershey-kiss,tab:hi-chew-bin,tab:hi-chew-ziploc} below, we provide a verbal description of the behavior that the expert demonstrations perform for each task. We additionally include a virtual rendering of different segments of a sample demonstration (where the gripper is rendered with increasing opacity for later timesteps). Additionally, we show a timelapse of the overhead camera image observation for the same sample expert demonstration.

\bgroup
\def\arraystretch{2}
\begin{table*}[p]
    \centering
    \begin{tabular}{m{2.4cm} m{10cm}} \toprule
        Task Name & \hiChewTitle (\hiChew) \\ 
        Task Description & \hiChewDescription \\
        Expert Trajectory Rendering & \includegraphics[width=\linewidth]{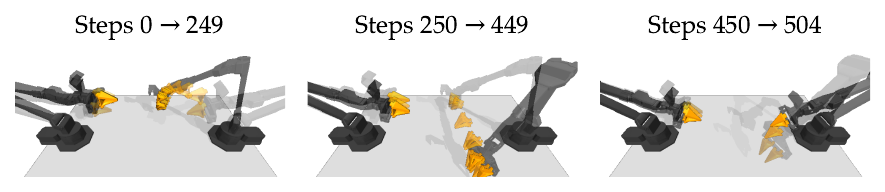} \\
        Expert Trajectory Timelapse & \includegraphics[width=\linewidth]{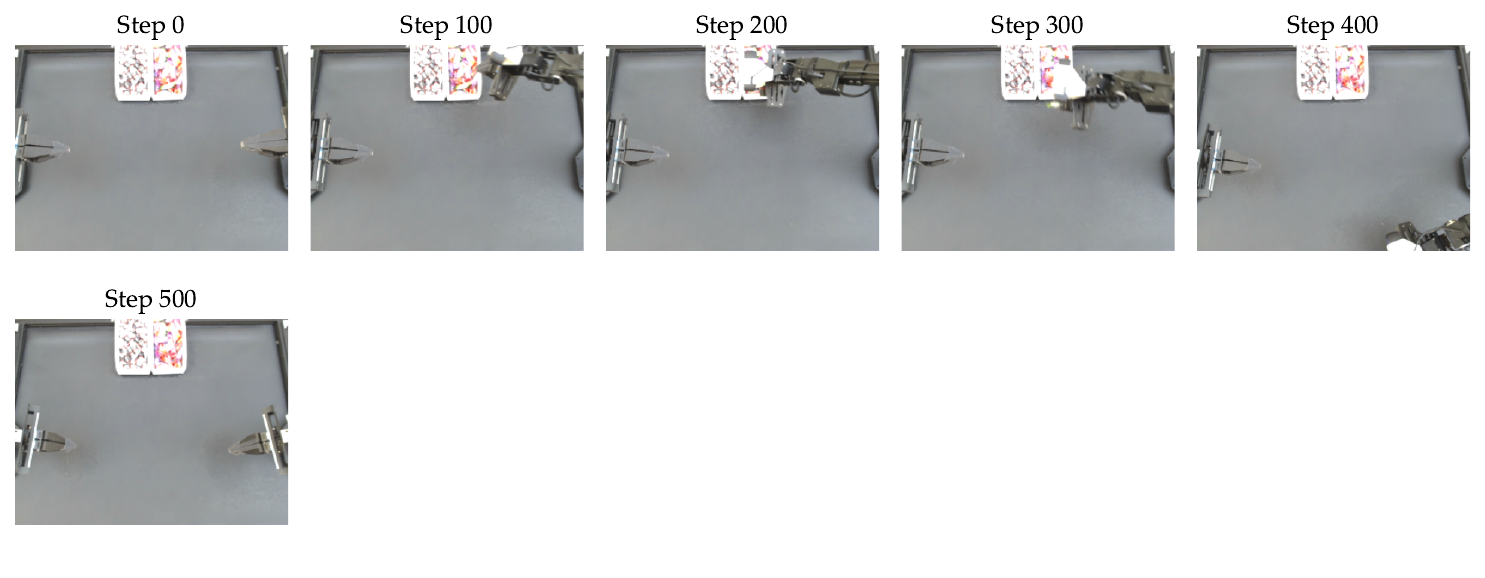}\\ \bottomrule
    \end{tabular}
    \caption{Description of the \hiChew task, as well as a rendering and timelapse of a sample expert trajectory.}
    \label{tab:hi-chew}
\end{table*}

\begin{table*}[p]
    \centering
    \begin{tabular}{m{2.4cm} m{10cm}} \toprule
        Task Name & \tootsieRollTitle (\tootsieRoll) \\
        Task Description & \tootsieRollDescription \\
        Expert Trajectory Rendering & \includegraphics[width=\linewidth]{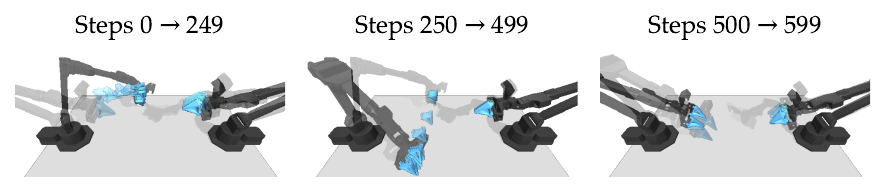} \\
        Expert Trajectory Timelapse & \includegraphics[width=\linewidth]{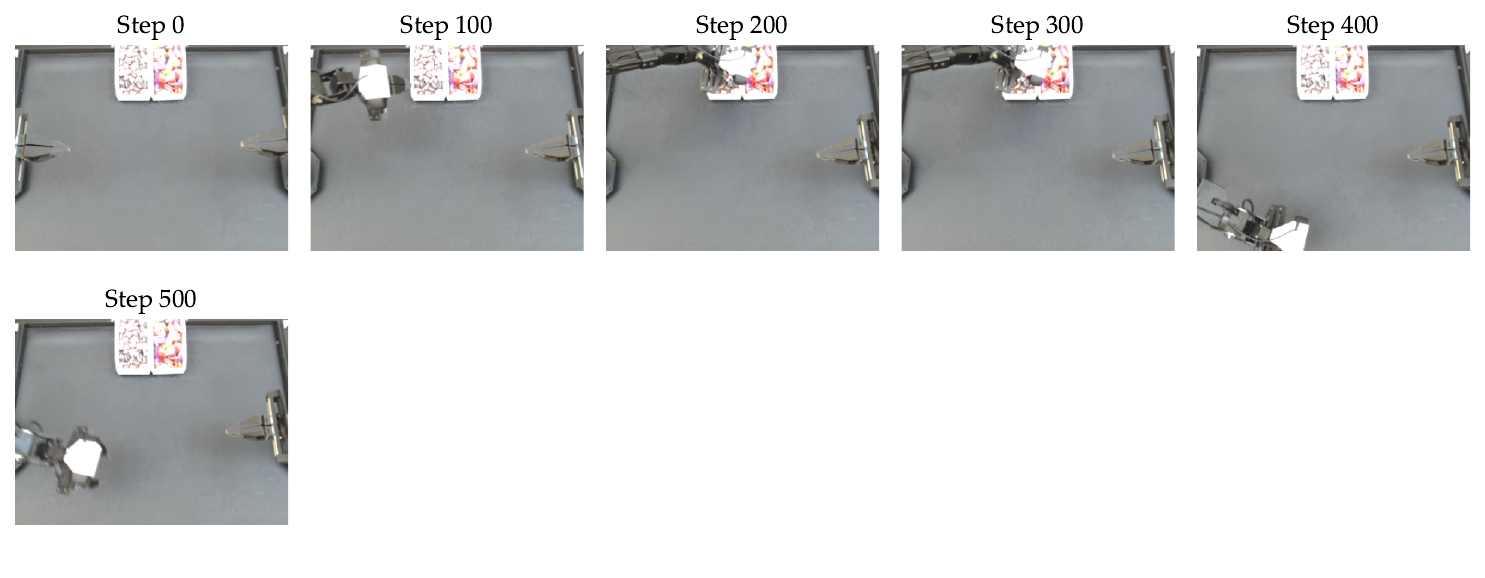}\\ \bottomrule
    \end{tabular}
    \caption{Description of the \tootsieRoll task, as well as a rendering and timelapse of a sample expert trajectory.}
    \label{tab:tootsie-roll}
\end{table*}

\begin{table*}[p]
    \centering
    \begin{tabular}{m{2.4cm} m{10cm}} \toprule
        Task Name & \hersheyKissTitle (\hersheyKiss) \\
        Task Description & \hersheyKissDescription \\
        Expert Trajectory Rendering & \includegraphics[width=\linewidth]{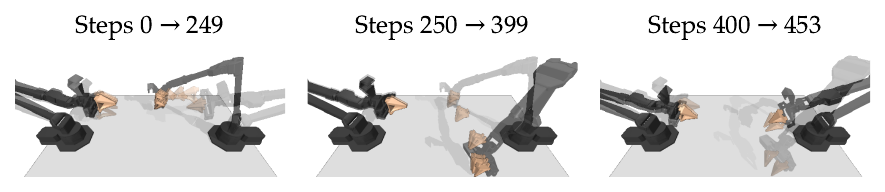} \\
        Expert Trajectory Timelapse & \includegraphics[width=\linewidth]{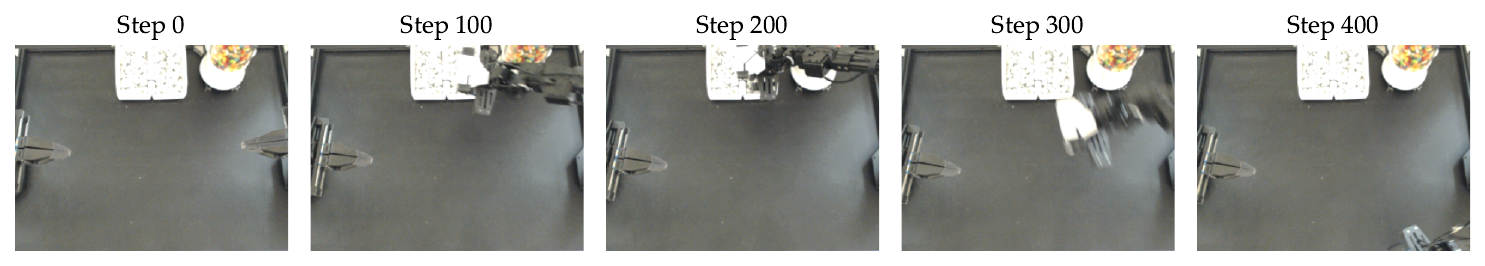}\\ \bottomrule
    \end{tabular}
    \caption{Description of the \hersheyKiss task, as well as a rendering and timelapse of a sample expert trajectory.}
    \label{tab:hershey-kiss}
\end{table*}

\begin{table*}[p]
    \centering
    \begin{tabular}{m{2.4cm} m{10cm}} \toprule
        Task Name & \jellyBeanTitle (\jellyBean) \\
        Task Description & \jellyBeanDescription \\
        Expert Trajectory Rendering & \includegraphics[width=\linewidth]{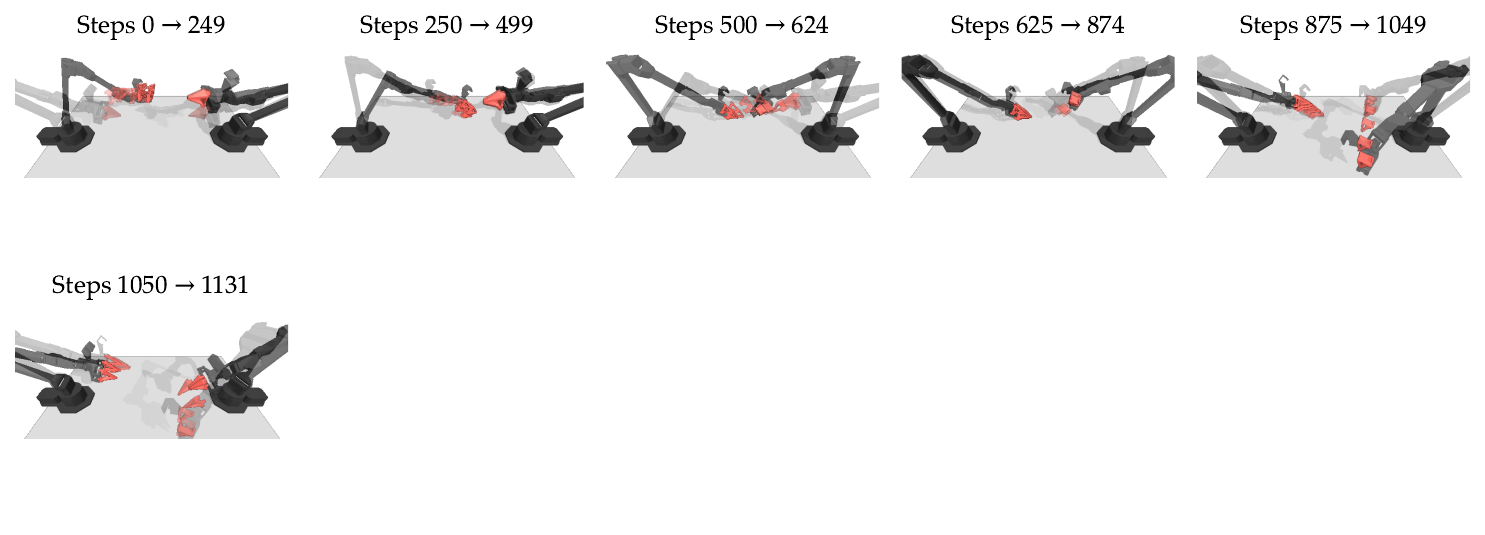} \\
        Expert Trajectory Timelapse & \includegraphics[width=\linewidth]{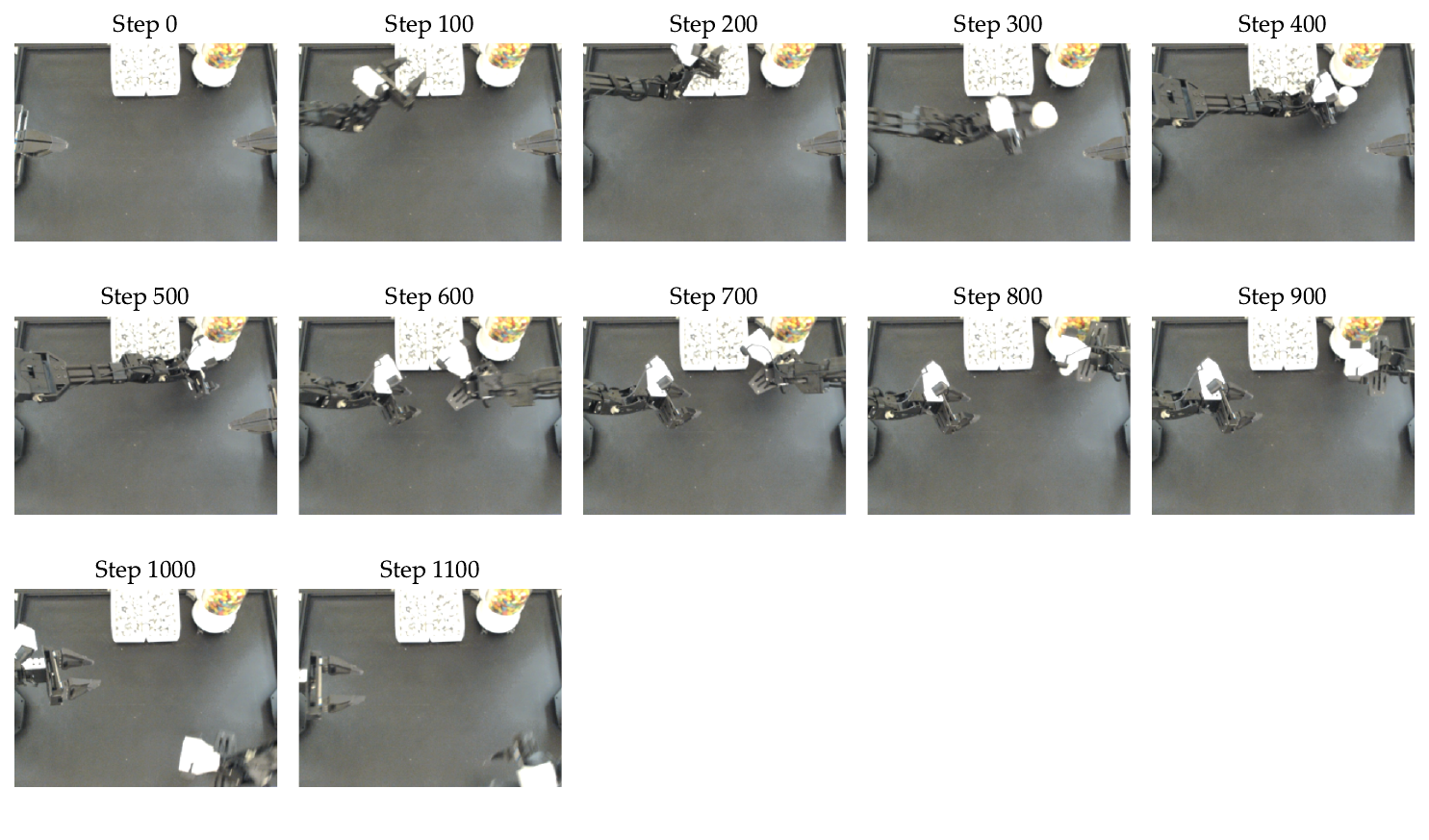}\\ \bottomrule
    \end{tabular}
    \caption{Description of the \jellyBean task, as well as a rendering and timelapse of a sample expert trajectory.}
    \label{tab:jelly-bean}
\end{table*}

\begin{table*}[p]
    \centering
    \begin{tabular}{m{2.4cm} m{10cm}} \toprule
        Task Name & \hiChewBinTitle (\hiChewBin) \\
        Task Description & \hiChewBinDescription \\
        Expert Trajectory Rendering & \includegraphics[width=\linewidth]{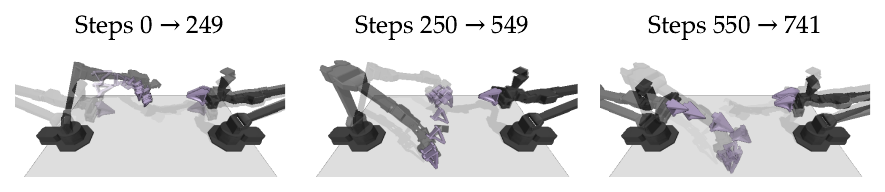} \\
        Expert Trajectory Timelapse & \includegraphics[width=\linewidth]{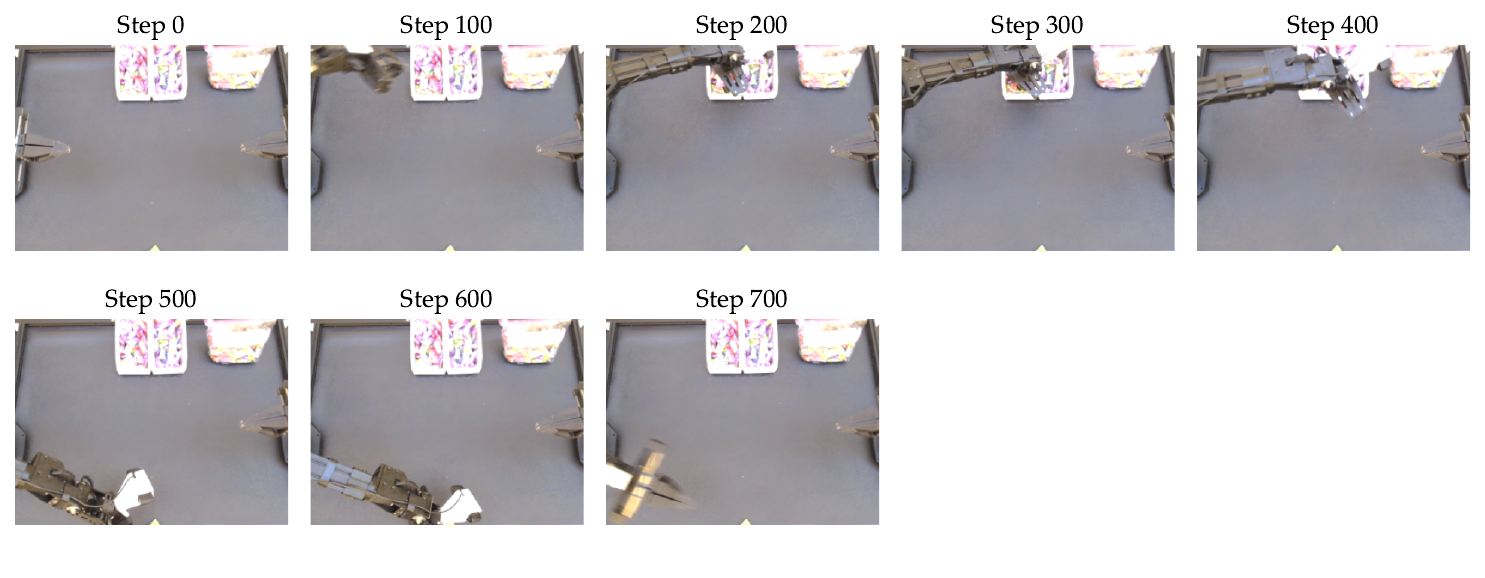}\\ \bottomrule
    \end{tabular}
    \caption{Description of the \hiChewBin task, as well as a rendering and timelapse of a sample expert trajectory.}
    \label{tab:hi-chew-bin}
\end{table*}

\begin{table*}[p]
    \centering
    \begin{tabular}{m{2.4cm} m{10cm}} \toprule
        Task Name & \hiChewZiplocTitle (\hiChewZiploc) \\
        Task Description & \hiChewZiplocDescription \\
        Expert Trajectory Rendering & \includegraphics[width=\linewidth]{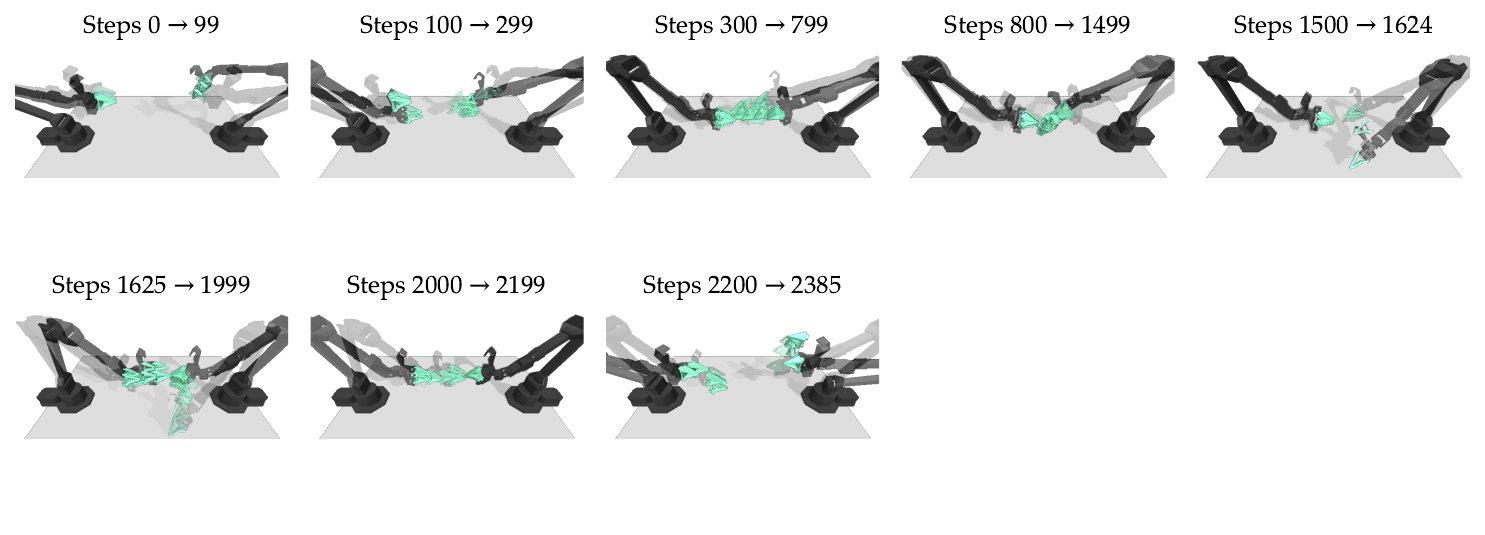} \\
        Expert Trajectory Timelapse & \includegraphics[width=\linewidth]{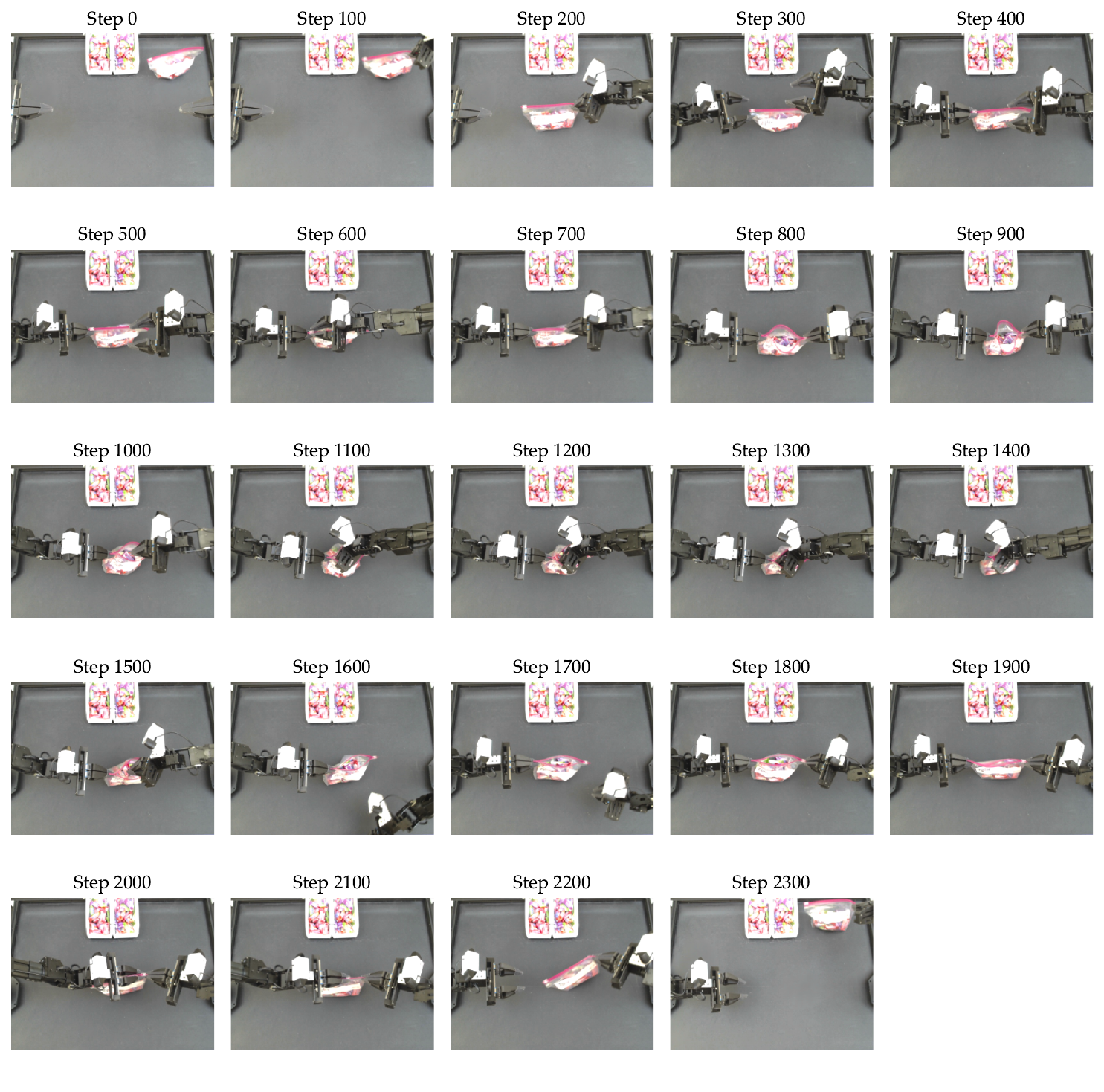}\\ \bottomrule
    \end{tabular}
    \caption{Description of the \hiChewZiploc task, as well as a rendering and timelapse of a sample expert trajectory.}
    \label{tab:hi-chew-ziploc}
\end{table*}

\egroup

\section{Dataset Examples}
\label{appx:dataset}
\begin{figure*}[p]
    \centering
    \includegraphics[width=0.7\linewidth]{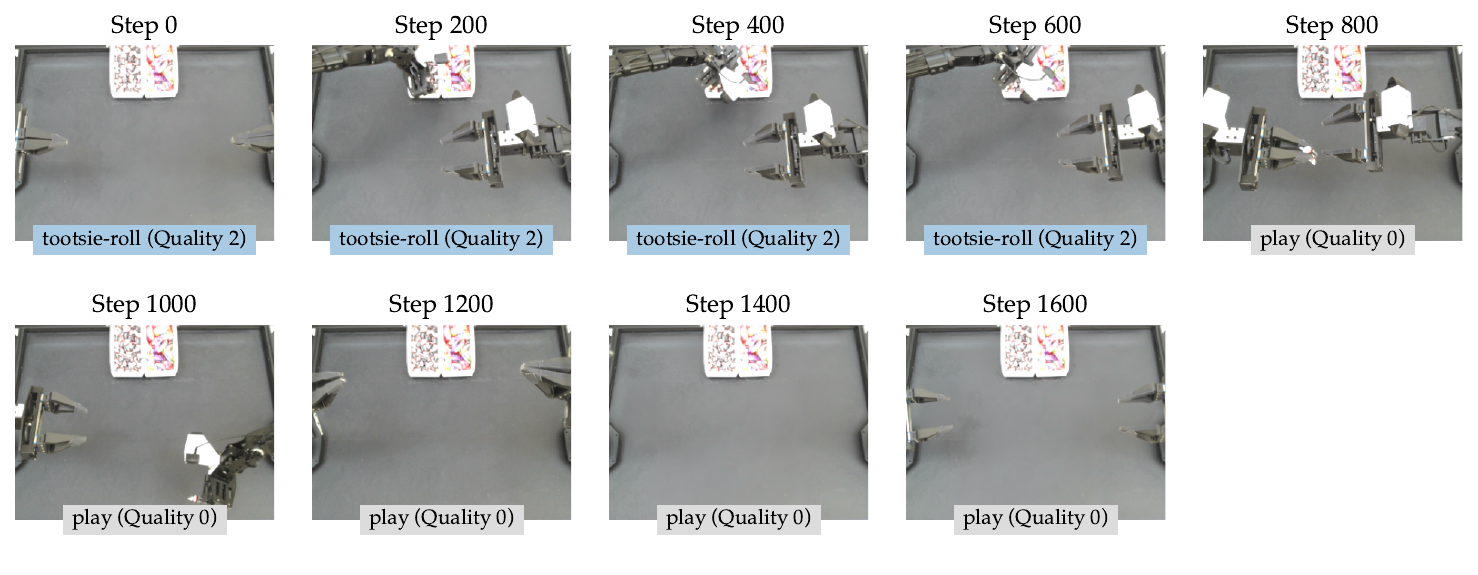}
    \caption{In this trajectory, the user begins by performing the \tootsieRoll task with moderate quality---i.e., there are about 3 attempts to grasp the candy, and there is some extraneous movement in the right arm, but the user is otherwise successful at grasping the candy. Before bringing the candy all the way to the End Zone, the user attempts to unwrap the candy. They then hand it over to the other arm, place it in the End Zone, and then move the arms upward. The first half of the episode is marked as \tootsieRoll (Quality 2) and the latter half of the episode is marked as \play (Quality 0).}
    \label{fig:crowd-1}
\end{figure*}

\begin{figure*}[p]
    \centering
    \includegraphics[width=0.7\linewidth]{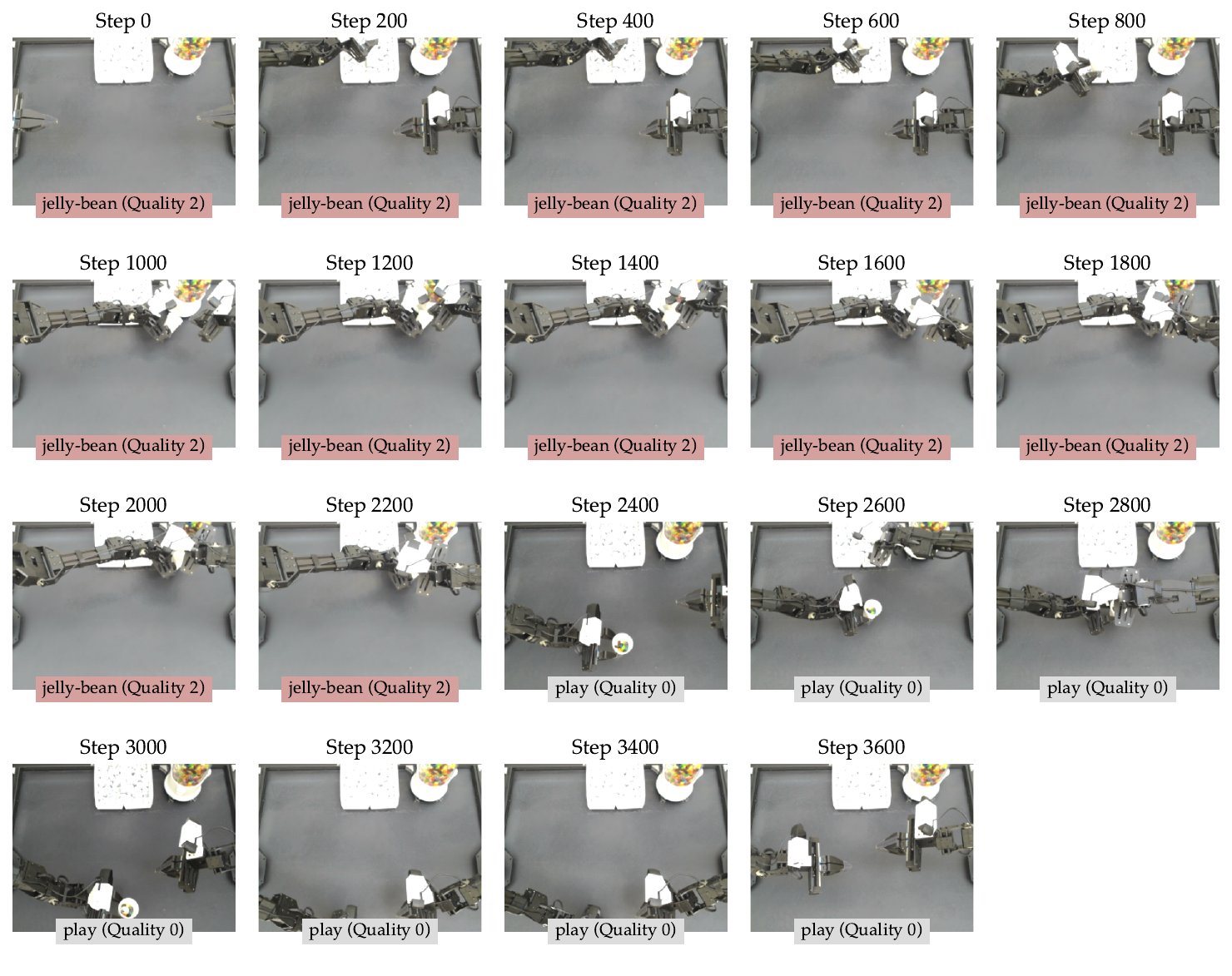}
    \caption{In this trajectory, the user grasps a cup from the cup dispenser and places it under the lever of the candy machine. They are successful in collecting jelly beans in the cup, though the trajectory includes retrying behavior and is not as smooth as an expert trajectory. The user brings the cup halfway to the End Zone, and then begins behaviors that are not part of the task---i.e., placing a Hershey Kiss in the cup before bringing it to the End Zone. The first part of the episode is marked as \jellyBean (Quality 2) and the latter part is marked as \play (Quality 0).}
    \label{fig:crowd-2}
\end{figure*}

\begin{figure*}[p]
    \centering
    \includegraphics[width=0.7\linewidth]{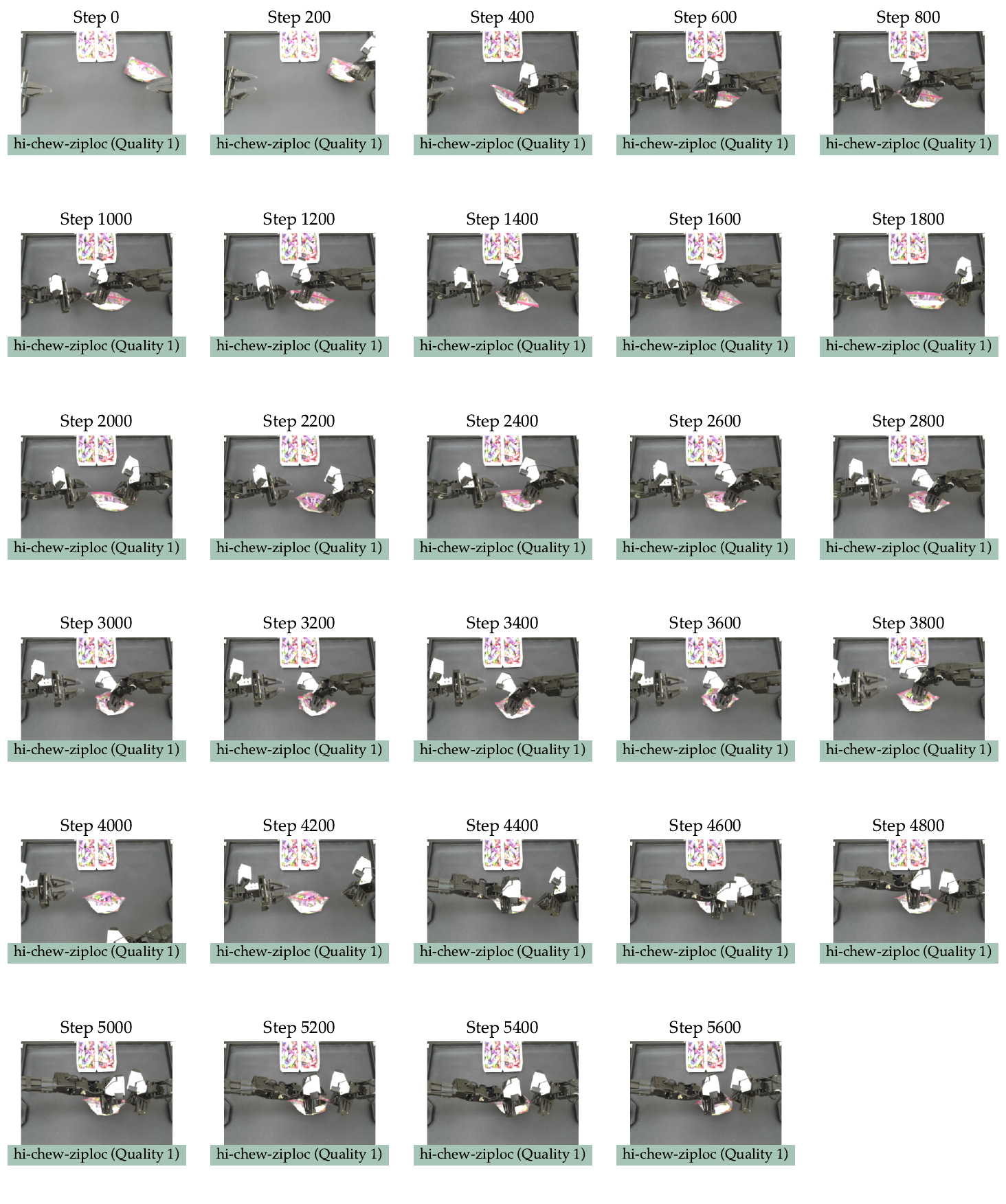}
    \caption{In this trajectory, the user correctly moves the Ziploc from the corner of the table to the center of the table, and grasps a Hi-Chew from inside the Ziploc which they bring to the End Zone. They are unsuccessful in closing the Ziploc before episode termination. The user is task-directed for the whole episode, however takes longer than better quality trajectories for this task and performs retrying behavior at each subtask. The whole trajectory is marked as \hiChewZiploc (Quality 1).}
    \label{fig:crowd-3}
\end{figure*}

In \cref{fig:crowd-1,fig:crowd-2,fig:crowd-3}, we give 3 qualitative examples of interaction episodes in our crowdsourced dataset. We illustrate a timelapse of each episode with the overhead camera observation. We also include the task and quality annotations at each timestep, with a verbal description of the episode in the caption.

\section{Additional Dataset Analysis}
\label{appx:dataset_analysis}
In this section, we provide additional data analysis. In \cref{appx:justification-for-scene}, we describe an offline study over user preferences for different candies, informing our different scene setups. In \cref{appx:additional-quality} and \cref{appx:likert}, we examine additional metrics (i.e., tutorial quality and Likert ratings) that correlate with quality of user interaction episodes, and in \cref{appx:usage}, we provide additional statistics on usage and retention.

\subsection{Justification for Scene Choices}
\label{appx:justification-for-scene}

To justify our scene setup and task pairings, we perform an offline survey on user preferences for various candies. On a sample of $N=16$ users, we find that $81\%$ prefer a Hi-Chew to a Tootsie Roll. Thus, \sceneA (which includes the \hiChew and \tootsieRoll tasks) allows us investigate whether this preference for material reward shapes task choice when teleoperating demonstrations, when the task is otherwise equivalent besides the material reward. Users exhibit a more mild preference for a Hershey Kiss compared to a small handful of Jelly Beans (with $62\%$ of respondents preferring the Hershey Kiss). \sceneC (which includes the \hiChewBin and \hiChewZiploc tasks) allows us to investigate how intrinsic motivation and task difficulty affects user behavior when teleoperating in the case that the material reward (a Hi-Chew) is held constant between the the simpler task and the more challenging task. \sceneB allows us to investigate this question when the material rewards are different, and users do not exhibit an overall preference for the reward from the harder task (and even mildly prefer the reward from the easier task).

\subsection{Additional Metrics on Demonstration Quality}
\label{appx:additional-quality}

Our crowdsourced dataset contains rich interaction data per user ID----during and after the interactive tutorial period. This dataset can help to yield insights about which users give higher quality trajectories, and what factors can help predict this quality. As an example, we examine how the quality of interactions \textit{after} the tutorial (i.e., when the user selects tasks in the scene to perform) correlates with quality \textit{during} the tutorial period (i.e., when the user is instructed to complete simple onboarding tasks). Specifically, we examine the distribution of mean quality during task interactions versus minimum quality during the tutorial period; the user's tutorial period is classified as 0 if there is any off-task behavior, 1 if the tutorial is performed but with retrying, and 2 if the tutorial is performed smoothly. We observe a loose positive correlation between higher minimum tutorial quality and mean task quality; and notably, users who produce consistently high quality task demonstrations (quality 3) are more present in the group with high quality tutorials. The tutorial period can therefore be a first-cut proxy at filtering demonstrators by quality.

\clearpage

\begin{figure}[ht]
    \centering
    \includegraphics[width=\linewidth]{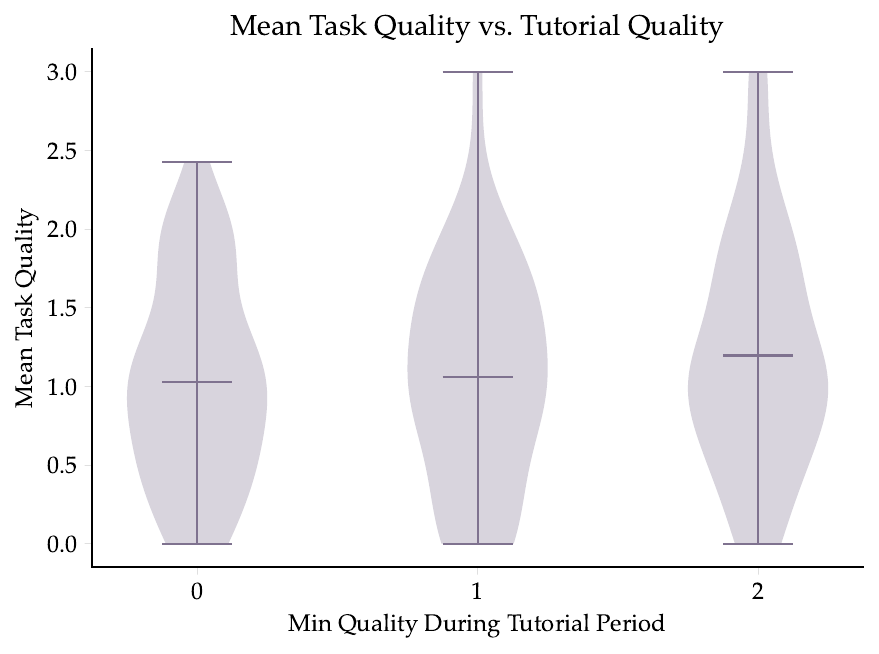}
    \caption{Distribution of Mean Task Quality versus Minimum Quality during the Tutorial Period.}
    \label{fig:tutorial-quality}
\end{figure}

\subsubsection{Self-Reported Likert Metrics}
\label{appx:likert}

\begin{figure*}[t]
    \centering
    \includegraphics[width=\linewidth]{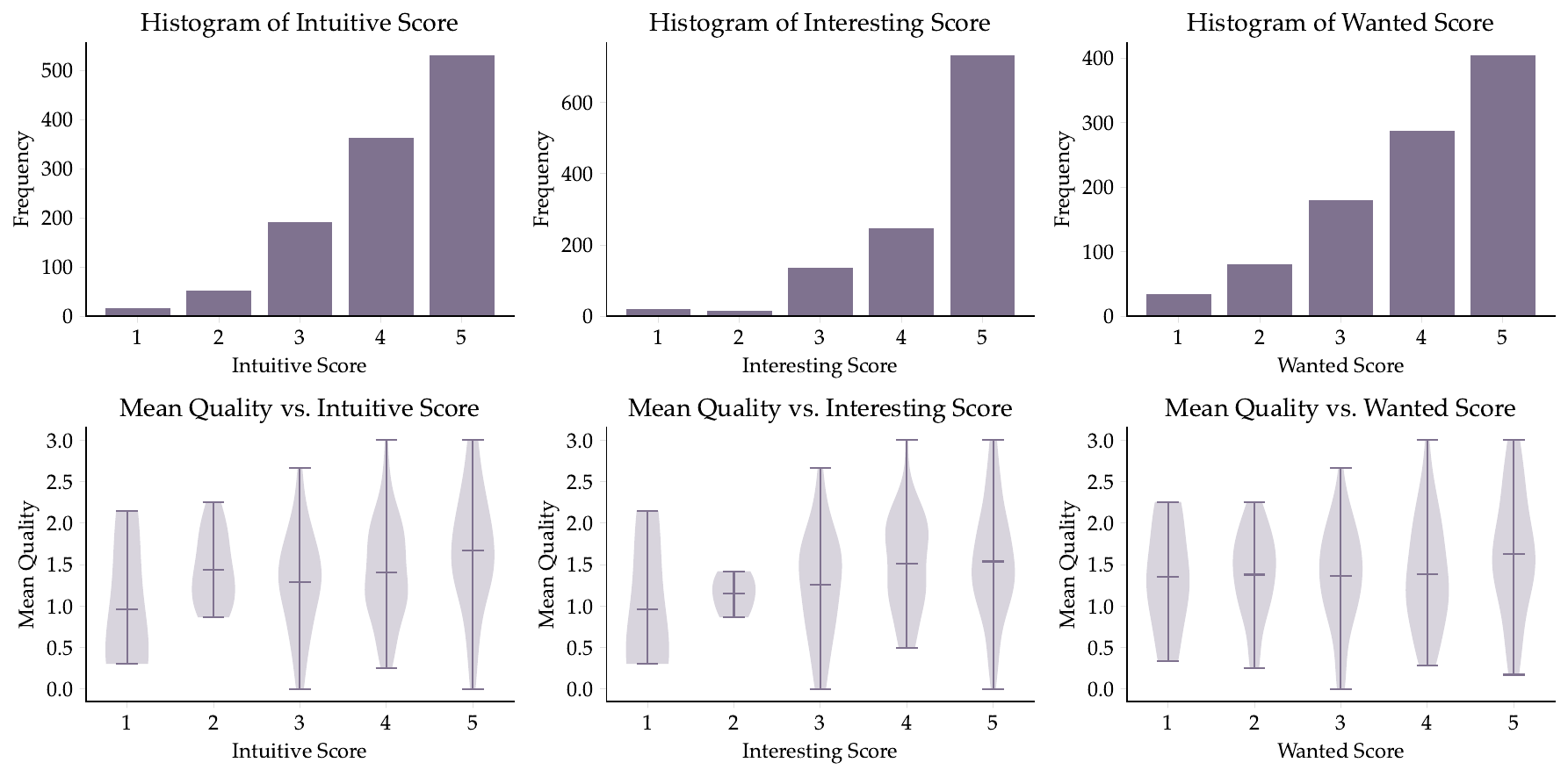}
    \caption{(\textit{Top}) Histogram of Likert Ratings (aggregated by the user's minimum response over their interaction episodes) for the \intuitive, \interesting, and \wanted questions. (\textit{Bottom}) Distribution of mean quality of interaction episodes for different Likert Ratings for \intuitive, \interesting, and \wanted.}
    \label{fig:likert}
\end{figure*}

After every interaction episode, we prompt the user to answer whether they agree with 3 statements, on a 5-point scale (1 - Strongly Disagree; 2 - Disagree; 3 - Neutral; 4 - Agree; 5 - Strongly Agree).
\begin{itemize}[leftmargin=*]
    \item \intuitive: Controlling the robot was intuitive.
    \item \interesting: Controlling the robot was fun and interesting.
    \item \wanted: The robot accomplished the task in the way that I wanted.
\end{itemize}
\cref{fig:likert} summarizes the responses to these questions, aggregated by users' minimum ratings to each statement over their interaction episodes. The majority of users agree with all three statements, and most often have the strongest ratings for \interesting compared to \intuitive and \wanted. We find also that there are loose correlations between the manually annotated quality scores for users' interaction episodes and users' self-reported ratings for each of these metrics. Specifically, users who self-report low ratings on each of the three metrics have lower mean quality scores. However, users who self-report high ratings have quality scores that span low to high.

\subsection{Usage and Retention}
\label{appx:usage}
\begin{figure*}[t]
    \centering
    \includegraphics[width=\linewidth]{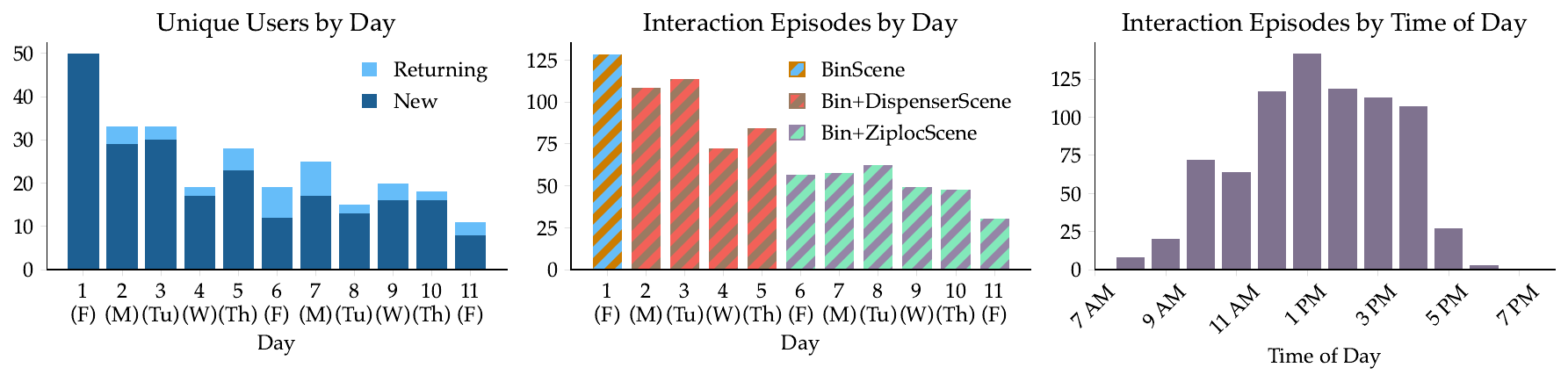}
    \caption{Statistics on usage over a two-week period: number of users per day (left), number of interaction episodes per day (middle), and distribution of interaction episodes by time of day (right).}
    \label{fig:usage}
\end{figure*}

We illustrate the usage of the \system in \cref{fig:usage}.
We observe significant engagement with \system over the two-week collection period: there were $N = $ 231 unique users in total.
On most days, more than two-thirds of these were new users that had not used the system on prior days. There were a total of 817 interaction episodes distributed throughout the period.
The most common time at which users interacted with the system was about 1pm, corresponding to the most trafficked time in the caf\'e (lunchtime). We collect 129 interaction episodes in \sceneA (Day 1), 381 in \sceneB (Days 2-5), and 307 in \sceneC (Days 6-11).

\section{Additional Details on Policy Learning Experiments}
\label{appx:policy_learning}
In this section, we give additional details on our policy learning experiments. \cref{appx:training-details} provides training details and hyperparameters, \cref{appx:evaluation-details} provides details on our evaluation procedure, and \cref{appx:qualitiative-analysis-policies} provides additional qualitative discussion of our learned policies.

\subsection{Training Details}
\label{appx:training-details}

For the \textit{Expert} and \textit{Co-train} experiments, we train policies for 200K steps for all tasks. For the \textit{Fine-tune} experiments, we fine-tune the co-trained model (partially trained for 100K steps) for an additional 50K steps on expert data only. We use the implementation of ACT \cite{Zhao2023LearningFB} from \cite{Fu2024MobileAL}, including the default hyperparameters from \cite{Zhao2023LearningFB}, as shown in \cref{tab:act-hyperparameters}.

\bgroup
\def\arraystretch{1.3}
\begin{table}[t]
    \small
    \centering
    \begin{tabular}{p{3cm}p{3cm}}
    \hline
    Learning Rate & 1e-5 \\
    Batch Size & 8 \\
    \# Encoder Layers & 4 \\
    \# Decoder Layers & 7 \\
    Feedforward Dimension & 3200 \\
    Hidden Dimension & 512 \\
    \# Heads & 8 \\
    Chunk Size & 100 \\
    KL-weight ($\beta$) & 10 \\
    Dropout & 0.1 \\
    Backbone & ResNet-18 \\
    Image Augmentations & RandomCrop, RandomResize, RandomRotation, ColorJitter \\
    \hline \\
    \end{tabular}
    \caption{Hyperparameters for ACT, shared for all experiments.}
    \label{tab:act-hyperparameters}
\end{table}
\egroup

\subsection{Evaluation Details}
\label{appx:evaluation-details}

We perform policy evaluations for 40 trials each, early stopping when policies exhibit excessively jittery or unsafe behavior. While the RoboCrowd training dataset was collected in a caf\'e where lighting varies throughout the day, during evaluation, we move the setup to a location with a visually similar background but consistent lighting for controlled evaluations.

For the bin-picking tasks, we define success as the robot arm picking exactly one of the desired candy and bringing it to the End Zone. For our challenging, long-horizon tasks (\jellyBean and \hiChewZiploc), success is $0\%$ for all policies, so we instead compare policies via  normalized return to measure partial proficiency at tasks. We describe the process for computing normalized return below.

Each of the following subtasks in \jellyBean corresponds to 1 point in the episode return: Retrieves Cup from Dispenser; Places Cup Down; Aligns Cup Under Lever; Presses Lever; Collects Jelly Beans in Cup; Picks up Cup; Brings Cup to End Zone. Each of the following subtasks in \hiChewZiploc corresponds to 1 point in the episode return: Picks up Bag; Places Bag in Center of Table; Slides Open; Picks Hi-Chew; Brings Hi-Chew to End Zone; Closes Bag; Places Bag in Corner of Table. For these tasks, we report normalized return---the average return over evaluation trials divided by the maximum return (achieved by all expert demonstrations).

\subsection{Qualitative Analysis of Learned Policies}
\label{appx:qualitiative-analysis-policies}

We find that in most cases, \textit{Co-train} and/or \textit{Fine-tune} improve upon \textit{Expert}. However, the specific effects vary by task. For example, we find that for the \hiChew task, the co-trained policy performs worse than the expert policy, but the fine-tuned policy performs better; whereas with the \hersheyKiss task, both the co-trained policy and fine-tuned policy perform better. We hypothesize that the crowdsourced data is more useful for \hersheyKiss because (a) \hersheyKiss is a more complex task (in that it is more multimodal, i.e., either arm can be used to pick up a Hershey Kiss, and the grasping required needs to be more precise to not crush the Hershey Kiss) and (b) a greater proportion of the \hersheyKiss data is of higher quality. We notice that the crowdsourced data for \jellyBean is especially diverse, and na\"ively co-training or fine-tuning underperforms using the expert data only.

Qualitatively, we observe in several cases that the co-trained and fine-tune policies exhibit meaningful but suboptimal behaviors from the crowdsourced data (e.g., picking up multiple objects from the bin instead of one). On the other hand, there are also helpful behaviors from the crowdsourced data (\textit{not} represented in the expert data) that benefit trained policies---e.g., regrasping behavior.

Overall, the RoboCrowd dataset is very diverse, and contains both task-relevant behaviors (of various levels of quality) and free-play behavior. Future work on more sophisticated policy learning methods that leverage these diverse characteristics can help to get the maximum utility out of crowdsourced demonstration data.

\section{Additional Details on Software Implementation and Data Annotation}
\label{appx:software}
In this section, we provide additional details on our software interface and implementation, as well as our data annotation pipeline. \cref{appx:application-flow} provides an overview of the application flow and interface, \cref{appx:interactive-tutorial} details the interactive tutorial procedure, \cref{appx:implementation-details} provides implementation details, and \cref{appx:data-annotation-pipeline} details the data annotation pipeline.

\subsection{Application Flow and User Interface}
\label{appx:application-flow}

\begin{figure}[htp]
    \centering
    \includegraphics[page=1, width=\linewidth, trim={0 7cm 0 0}, clip]{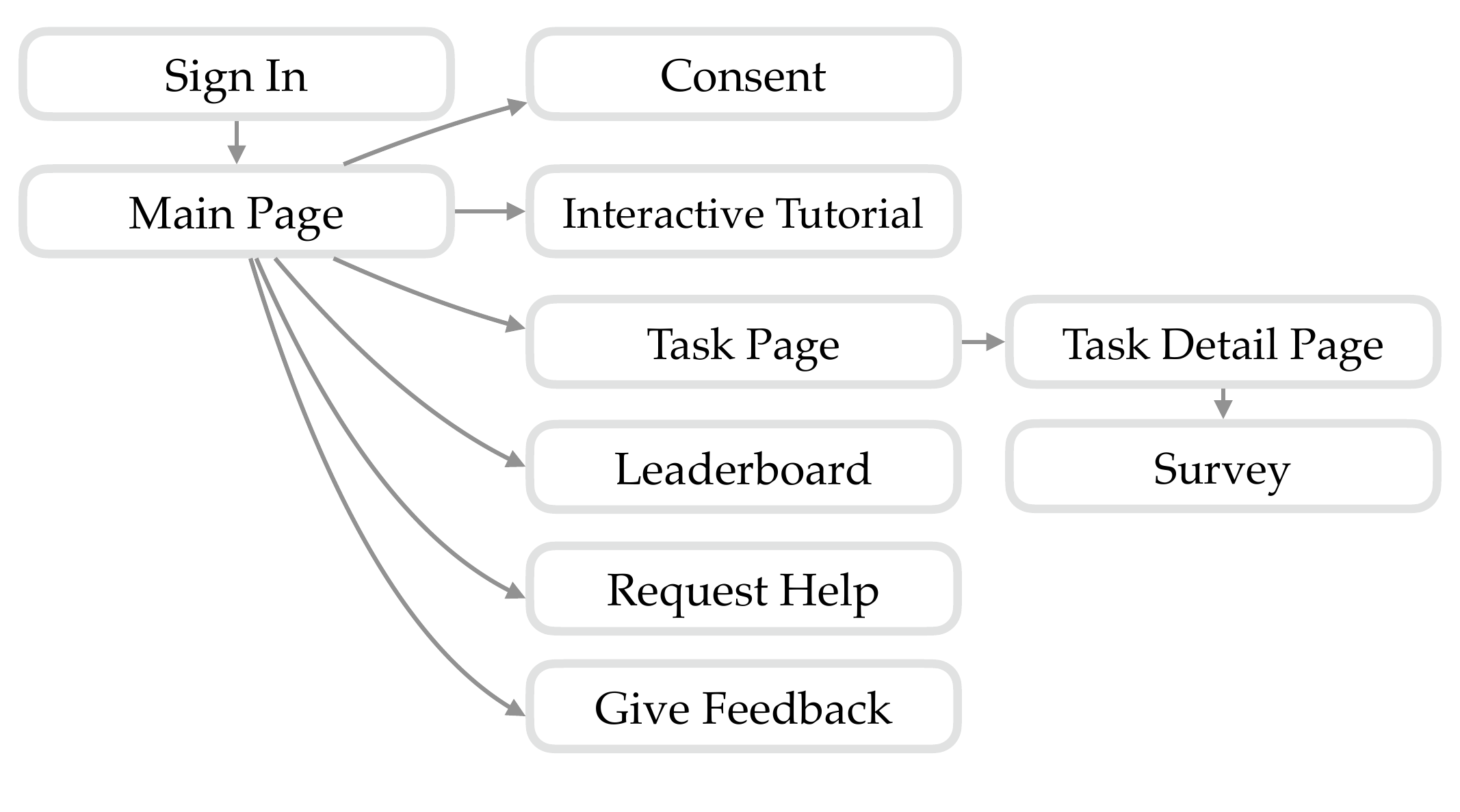}
    \caption{Flowchart illustration of pages in the user interface.}
    \label{fig:interface-full}
\end{figure}

\cref{fig:interface-full} gives an overview of the flow through the tablet application, and \cref{tab:interface-examples} provides screenshots of the major pages referenced in the flowchart. We additionally highlight the Interactive Tutorial in \cref{fig:interactive-tutorial-pages} and the visual warning for collision detection in \cref{fig:collision-warning}. We now briefly describe the application flow. To begin a new session, the user taps their ID card on the card reader, which advances the tablet application to a screen where the user can enter a nickname (if they are a new user). They are then directed to the Main Page, where they complete a consent form and the interactive tutorial. From the Main Page, users can also press a ``Start Playing'' button which directs them to the Task Page, where they can see videos of tasks available in the scene, and can tap on a task to see more details and begin demonstrating the task. For safety, the user receives an audial and visual warning (\cref{fig:collision-warning}) if the arms are near-collision. When users are done with the task (i.e., they click a Stop button on the Task Detail Page or they rest the grippers on the mechanical stop), they are asked to mark their demonstration as a success or failure, and fill out a brief survey. The success/failure markings are used as the basis for the points which are added to the user's point total in the Leaderboard, which is accessible from the Main Page; in our experiments, users receive 10 points for successful ``easy'' tasks (bin-picking) and 20 points for successful ``difficult'' tasks (the remaining tasks). From the Main Page, users can also choose to provide feedback, or press a Request Help button which immediately notifies the study team (e.g., if the user needs assistance or if the setup requires maintenance).

\subsection{Interactive Tutorial}
\label{appx:interactive-tutorial}

We provide a zoomed-in version of the pages in the Interactive Tutorial in \cref{fig:interactive-tutorial-pages}. The aim of the tutorial is to guide the user on how to start and stop interaction episodes as well as how to puppeteer with ALOHA. Specifically, users are first instructed to wait until ALOHA's arms rise to the home position, and then they are given instructions on how to start puppeteering (by squeezing both sets of grippers on the leader arms). After they do so, the tutorial automatically proceeds to the next stage, where users then are told to gently touch the left and right arms to the table; the goal is to help users get calibrated to the robot's range of motion and degrees of freedom, as well as the types of forces they need to apply to move the arms. Finally, users are given instructions on how to stop the interaction episode, by resting the grippers of the leader arms in the grooves of the mechanical stops. When the user does so, the puppet arms are automatically lowered, and the user is presented a brief video on how to navigate the rest of the interface.

\subsection{Implementation Details}
\label{appx:implementation-details}

The software application is implemented with React (frontend) and Flask (backend), and uses WebSocket connections to communicate between the user client and backend server. We use a SocketIO-ROS bridge to pass messages between the backend server and robot controller. The robot controller operates at 50Hz and is based on \cite{Zhao2023LearningFB}. When the robot is being teleoperated, we run a parallel simulation in MuJoCo \cite{todorov2012mujoco} which is updated at every time step to detect self-collisions. 

\begin{figure*}[htp]
    \centering
    \includegraphics[page=5, width=0.9\linewidth, trim={0 6cm 0 0}]{figures-appendix/interface_v2.pdf}
    \caption{Screenshot of the pages in the interactive tutorial interface.}
    \label{fig:interactive-tutorial-pages}
\end{figure*}

\begin{figure}[htp]
    \centering
    \includegraphics[page=12, width=\linewidth, trim={0 0cm 0.01cm 0}]{figures-appendix/interface_v2.pdf}
    \caption{Screenshot of a visual collision warning on the task page. An audial alarm (beeping sound) is played on the tablet when the visual collision warning appears.}
    \label{fig:collision-warning}
\end{figure}

\bgroup
\def\arraystretch{1.5}
\begin{table*}[p]
    \centering
    \begin{tabular}{m{2.5cm}m{4.5cm}m{0.5cm}m{2.5cm} m{4.5cm}} \toprule
        Page Name & Screenshot && Page Name & Screenshot \\ \midrule \\
        Sign In (Tap ID Card) & \includegraphics[page=2, width=\linewidth, trim={0 3cm 0 0}]{figures-appendix/interface_v2.pdf} &&
        Sign In (Create User Profile) & \includegraphics[page=3, width=\linewidth, trim={0 3cm 0 0}]{figures-appendix/interface_v2.pdf} \\
        Main Page & \includegraphics[page=4, width=\linewidth, trim={0 3cm 0 0}]{figures-appendix/interface_v2.pdf} &&
        Interactive Tutorial & \includegraphics[page=5, width=\linewidth, trim={0 3cm 0 0}]{figures-appendix/interface_v2.pdf} \\
        Task Page & \includegraphics[page=6, width=\linewidth, trim={0 3cm 0 0}]{figures-appendix/interface_v2.pdf} &&
        Task Detail Page & \includegraphics[page=7, width=\linewidth, trim={0 3cm 0 0}]{figures-appendix/interface_v2.pdf} \\
        Leaderboard & \includegraphics[page=9, width=\linewidth, trim={0 3cm 0 0}]{figures-appendix/interface_v2.pdf} &&
        Survey Page & \includegraphics[page=8, width=\linewidth, trim={0 3cm 0 0}]{figures-appendix/interface_v2.pdf} \\
        Request Help & \includegraphics[page=10, width=\linewidth, trim={0 3cm 0 0}]{figures-appendix/interface_v2.pdf} &&
        Give Feedback & \includegraphics[page=11, width=\linewidth, trim={0 3cm 0 0}]{figures-appendix/interface_v2.pdf} \\ \\ \bottomrule
    \end{tabular}
    \caption{Screenshots of pages in the user interface.}
    \label{tab:interface-examples}
\end{table*}
\egroup

\subsection{Data Annotation Pipeline}
\label{appx:data-annotation-pipeline}

We hand-annotate episodes in our crowdsourced dataset by task and quality. We implement an interface for annotation, which we illustrate in \cref{fig:annotation-interface}. We annotate episodes by dragging a slider which scrubs through the episode and selecting a task and quality annotation for different segments of the episode. We describe the annotation rules below.

\begin{itemize}
    \item \play (Quality 0). All free-play behavior is marked as \play with quality 0. Play data includes undirected movements and tasks that the user makes up (e.g., trying to unwrap a candy). It also includes extraneous movements before and after the user performs a task.
    \item \tutorial (Quality 1--2). Movements associated with the tutorial (e.g., touching the grippers to the table) are marked as Q1 if there is any retrying behavior and Q2 if the motions are smooth (i.e., lacking jerky or sudden movements).
    \item \texttt{<task>} (Quality 1--3). Task-relevant motions for each of our six tasks are labeled with the task name and a quality from 1 to 3. Q3 is used to describe segments that complete subtasks smoothly (without jerky motions) with no more than 2 retries. Q2 is used to describe segments that use no more than 4 retries for any one subtask, or that are completed but with slight errors (e.g., grabbing more than 1 candy from a bin). Q1 is used to describe segments that are task-relevant but of poor quality (e.g., more than 4 retries for any one subtask), cause changes to the scene (e.g., dropping a candy on the table), or complete the task in a significantly different manner than the expert demonstrations (e.g., using the opposite arm for any subtask). 
\end{itemize}

\begin{figure}[htp]
    \centering
    \includegraphics[page=13, width=\linewidth, trim={0 10cm 0 0}]{figures-appendix/interface_v2.pdf}
    \caption{Screenshot of the data annotation interface. Annotators can scrub through the episode and label segments with task and quality labels, which color codes a bar to visualize the different tasks and qualities in the episode. When the annotator is done labeling an episode, they can ``commit'' their labels and proceed to the next episode.}
    \label{fig:annotation-interface}
\end{figure}

\section{Additional Details on Pilot Studies and System Development}
\label{appx:pilots}
Prior to full system deployment, we conducted pilot studies on a smaller population to help us iterate on our system. We obtained the Institutional Review Board’s approval before both the pilot studies and the full deployment. We recruited $N$=10 participants to interact with the system. In order to mimic organic interactions as closely as possible, we did not provide the participants with any verbal instructions, other than to begin interacting with the system as if they happened upon it organically. Our software interface guided the participants through the consent form and tutorial. Here is a sample of feedback provided by participants, coupled with changes we made to the system.
\begin{itemize}[leftmargin=*]
    \item \textit{Degrees of Freedom}: Users indicated that puppeteering demonstrations was challenging the first time because they needed to ``understand the degrees of freedom'' of the robot. To address this feedback, we created a tutorial where the user was guided through how to perform primitive movements of the leader arms (e.g., controlling both puppet arms to touch the bottom of the workspace) before they began interacting with the system.
    \item \textit{Tutorial Format}: In an initial prototype, our tutorial was a video that a user would watch before using the system. Users provided feedback that they felt ``impatient'' and would rather ``explore what it is like to interface with the robot'' rather than ``watch a long video.'' To address this feedback, we made the tutorial efficient and interactive: 4 steps that the user would perform with the robot after watching them on the screen. The interactive tutorial automatically advances after detecting that each step is complete.
    \item \textit{Start and Stopping Demonstrations}: In an initial prototype, users begin demonstrations by (1) tapping a Start button on an interface and (2) squeezing the grippers of the leader arms closed. To terminate episodes, they would simply need to (1) leave the arms to rest on the robot body and (2) tap a Stop button on the interface. We received feedback that squeezing the gripper to start episodes ``made sense'' but the ``rest position at the end was confusing.'' To address this feedback, we designed and 3D printed a mechanical stop for users to rest the arms. We automatically terminate episodes when handles of the leader arms make contact with this mechanical stop.
    \item \textit{Interface}: In an initial prototype, users would access the interface on their own smartphone by scanning a QR code pasted on the platform. A user reported that they would prefer if more of their interaction would happen ``in the position that they will be doing the task.'' We therefore switched to a tablet interface mounted at the base of the platform, which was accessible when the user sat down to begin interacting with the robot. On the interface itself, users reported that it was ``easy to understand.''
    \item \textit{Collisions}: We observed that participants did not actively pay much attention to collisions between the robots, as well as the collision of wrist-camera mounts and objects mounted on the table. To address this, we (1) added collision avoidance between the arms and the table, (2) added an audio-visual alarm when arms were near collision, and (3) mounted objects to the table so that they would not move.
\end{itemize}

\section{Overview of Action Chunking with Transformers (ACT)}
\label{appx:act}
In this section, we provide a more extended background overview of imitation learning (IL) and the Action Chunking with Transformers (ACT) algorithm \citep{Zhao2023LearningFB}.

Imitation learning (IL) aims to learn a policy $\pi_\theta$ parameterized by $\theta$ given access to a dataset $\mathcal{D}$ composed of expert demonstrations. Defined within the framework of a standard partially observable Markov decision process (POMDP), each trajectory $\xi \in \mathcal{D}$ is a sequence of observation-action transitions $\{(o_0, a_0), \ldots, (o_T, a_T)\}$. Most commonly, IL is instantiated as behavior cloning, which trains $\pi_\theta$ to minimize the negative log-likelihood of data, $\mathcal{L}(\theta) = -\mathbb{E}_{(o,a)\sim \mathcal{D}}[\log \pi_\theta (a | o)]$.

In practice, the human-collected demonstrations in $\mathcal{D}$ may be diverse. To effectively learn from such diverse data, we can condition the policy on a latent variable $z$, which helps to capture the variability in the demonstrations by representing different modes of behavior. Representing this policy as the decoder in a conditional variational autoencoder (cVAE), we in addition learn an encoder $q_\phi$ from (observation, action) pairs to the latent space: $q_\phi(z \mid a^t,o^t)$. And we condition our policy on the latent variable: $\pi_\theta(\hat{a}^t \mid o_t,z)$. At test time, we sample latent vectors from the standard normal distribution, $z\sim\mathcal{N}(0,1)$. We regularize the outputs of our encoder towards this distribution via a KL-penalty: $D_{KL}(q_\phi(z \mid a^t,o^t) \mathrel{\Vert} \mathcal{N}(0,1))$. This method is formalized as Action Chunking with Transformers (ACT) \citep{Zhao2023LearningFB}, an imitation learning algorithm designed to learn from diverse human demonstrations.

\end{document}